%% file: main.tex
\documentclass[letterpaper]{article} % DO NOT CHANGE THIS
\usepackage{aaai25}  % DO NOT CHANGE THIS
\usepackage{times}  % DO NOT CHANGE THIS
\usepackage{helvet}  % DO NOT CHANGE THIS
\usepackage{courier}  % DO NOT CHANGE THIS
\usepackage[hyphens]{url}  % DO NOT CHANGE THIS
\usepackage{graphicx} % DO NOT CHANGE THIS
\usepackage{pdfpages}
\usepackage{natbib}  % DO NOT CHANGE THIS AND DO NOT ADD ANY OPTIONS TO IT
\usepackage{caption} % DO NOT CHANGE THIS AND DO NOT ADD ANY OPTIONS TO IT
\usepackage{siunitx}
\input{preamble}

\renewcommand{\first}{\textbf}
\renewcommand{\second}{\underline}
\usepackage{multirow}
\usepackage{booktabs}
\usepackage{pifont}
\usepackage{bm}
\usepackage{enumitem}
\usepackage{newfloat}
\usepackage{listings}

\DeclareCaptionStyle{ruled}{labelfont=normalfont,labelsep=colon,strut=off} % DO NOT CHANGE THIS
\floatstyle{ruled}
\newfloat{listing}{tb}{lst}{}
\floatname{listing}{Listing}

\title{Neural Variable-Order Fractional Differential Equation Networks}
\author{
    Wenjun Cui\equalcontrib \textsuperscript{\rm 1}\textsuperscript{\rm 3}, Qiyu Kang\equalcontrib \textsuperscript{\rm 2}, Xuhao Li\textsuperscript{\rm 4}, Kai Zhao\textsuperscript{\rm 5}, Wee Peng Tay\textsuperscript{\rm 5}, Weihua Deng\textsuperscript{\rm 6}, Yidong Li\thanks{Corresponding author (ydli@bjtu.edu.cn).}\textsuperscript{\rm 1}\textsuperscript{\rm 3}\\
}
\affiliations{
    \textsuperscript{\rm 1} Key Laboratory of Big Data \& Artificial Intelligence in Transportation (Beijing Jiaotong University)\\
    \textsuperscript{\rm 2} University of Science and Technology of China\\
    \textsuperscript{\rm 3} Beijing Jiaotong University\\
    \textsuperscript{\rm 4} Anhui University\\
    \textsuperscript{\rm 5} Nanyang Technological University\\ \textsuperscript{\rm 6} Lanzhou University\\
}

\usepackage{textgreek} % For text mode Greek letters
\newcommand{\altalpha}{\text{\textalpha}}

\usepackage{bibentry}
\begin{document}

\maketitle

\begin{abstract}
    Neural differential equation models have garnered significant attention in recent years for their effectiveness in machine learning applications.  
   Among these, fractional differential equations (FDEs) have emerged as a promising tool due to their ability to capture memory-dependent dynamics, which are often challenging to model with traditional integer-order approaches.
    While existing models have primarily focused on constant-order fractional derivatives, variable-order fractional operators offer a more flexible and expressive framework for modeling complex memory patterns. In this work, we introduce the Neural Variable-Order Fractional Differential Equation network (NvoFDE), a novel neural network framework that integrates variable-order fractional derivatives with learnable neural networks. 
Our framework allows for the modeling of adaptive derivative orders dependent on hidden features, capturing more complex feature-updating dynamics and providing enhanced flexibility. We conduct extensive experiments across multiple graph datasets to validate the effectiveness of our approach. Our results demonstrate that NvoFDE outperforms traditional constant-order fractional and integer models across a range of tasks, showcasing its superior adaptability and performance.
% The implementation code is available at
% \link{Code}{https://github.com/cuiwjTech/AAAI2025_NvoFDE}.

% The implementation code is available at \url{https://github.com/cuiwjTech/AAAI2025_NvoFDE}.
\end{abstract}


\section{Introduction}\label{sec.intro}
The intersection of differential equations and machine learning has opened a new frontier for developing algorithms that benefit both communities. 
Machine learning, for example, has been adeptly applied to solve high-dimensional partial differential equations (PDEs), particularly valuable in ill-posed and inverse problems where traditional numerical methods falter. 
Notable examples include PINNs and their variants \cite{raissi2019physics, zhu2019physics}, which leverage both measurement data and PDE constraints through automatic differentiation, as enabled by modern platforms like TensorFlow \cite{abadi2016tensorflow} and PyTorch \cite{paszke2019pytorch}. This progress has wide-ranging applications in areas such as quantum chemistry \cite{pfau2020ab}, materials science \cite{shukla2020physics}, geophysics \cite{zhu2021general}, and molecular simulations \cite{zhang2018deep}. 
Conversely, differential equations have enriched machine learning \cite{weinan2017proposal}, exemplified by integer-order neural ordinary differential equations (ODEs) \cite{chen2018neural} that model continuous residual layers. 
This integration has spurred enhancements in neural network performance \cite{dupont2019augmented,dai2024recode}, gradient stability \cite{haber2017stable,gravina2022anti}, and robustness \cite{yan2019robustness,kang2021Neurips,cui2023robustness}. Moreover, recent applications of stochastic differential equations in machine learning include generating data from noise, modeling intricate financial dynamics and enabling uncertainty quantification via Bayesian approaches \cite{songscore, xu2022infinitely}.
%In the mathematical literature, the differential operator $\frac{\mathrm{d}^\alpha}{\mathrm{d} t^\alpha}$ has been extended to accommodate non-integer values of $\alpha$, such as $\alpha=0.5$. 

Fractional differential operators have shown considerable promise in describing real-world phenomena more effectively than their integer-order counterparts. 
Unlike traditional integer-order differential equations, fractional differential equations (FDEs) provide a way to incorporate a continuum of past states into the present state, offering a rich and flexible modeling framework.
FDEs have found applications across diverse fields, including material science \cite{coleman1961foundations}, signal processing \cite{machado2011recent}, finance \cite{scalas2000fractional}, and porous and fractal phenomena modeling \cite{nigmatullin1986realization}. 
In machine learning, fractional derivatives have been used to optimize neural network parameters \cite{liu2022regularized}, diverging from the conventional integer-order derivatives employed in algorithms such as SGD or Adam \cite{kingma2014adam}.
Moreover, fractional calculus has been applied to enhance densely connected graph neural networks (GNNs), addressing issues like non-smooth data and the vanishing gradient problem \cite{antil2020fractional}. Recently, \cite{nobis2023generative} proposes generative fractional diffusion models that achieve greater pixel-wise diversity and improved image quality.  
Additionally, FDE-based GNNs, employing fractional diffusion and oscillator mechanisms to propagate information over graphs, have demonstrated superior capabilities in graph representation learning and task performance compared to their integer-order counterparts \cite{KanZhaDin:C24,ZhaKanSon:C24}.

Existing work on employing FDEs within machine learning has predominantly focused on \emph{constant-order} fractional derivatives, which assume the order remains fixed over time. This assumption often falls short of capturing the varying impact of historical information across different contexts and systems.
Recent advancements have introduced \emph{variable-order} fractional differential equations \cite{samko1993integration,coimbra2003mechanics}, which allow the order $\altalpha$ of the fractional-order derivative $\frac{\mathrm{d}^\altalpha}{\mathrm{d} t^\altalpha}$ to vary over time. This generalization has proved essential for modeling a range of real-world phenomena in science and engineering \cite{coimbra2003mechanics,obembe2017variable}.
% \cite{chechkin2005fractional,coimbra2003mechanics,obembe2017variable}.
In such systems, the order of the differential equation may depend on specific variables like temperature, concentration, or density. 
For example, \cite{glockle1995fractional} found the differential order governing protein relaxation varies with temperature changes. Similarly, \cite{meng2016fractional} illustrated that a variable-order fractional viscoelastic model more precisely describes the time-dependent evolution of mechanical properties in viscoelastic materials.
Work that combines variable-order FDEs with neural networks has been more related to computational neuroscience \cite{anastasio1994fractional} or Hopfield networks \cite{kaslik2012nonlinear}, focusing on numerical simulations and the analysis of bifurcation and stability behavior within connected networks \cite{yang2018stability,xu2019influence}. Additionally, researchers have applied neural networks to solve variable-order FDEs, with notable examples found in \cite{lasaki2023novel}.
% examples found in \cite{lasaki2023novel, zuniga2018new}.

Building upon these developments, our work introduces the Neural Variable-Order Fractional Differential Equation network (NvoFDE), a novel framework that integrates the flexibility of VoFDEs with the expressive power of learnable neural networks. 
In contrast to previous approaches like \cite{chen2018neural,KanZhaDin:C24} that utilize a constant derivative order $\altalpha$, NvoFDE allows the order to be of the form $\altalpha(t, \bx)$, which varies according to the time and hidden features. 
This adaptation offers a powerful tool for modeling non-uniform memory effects, enabling the neural network to adjust its sensitivity to preceding hidden features, which could enhance task performance.
To validate our framework, we conduct extensive experiments across multiple graph datasets and benchmark NvoFDE against traditional fixed-order models. Our results demonstrate that NvoFDE not only adapts more effectively to varying datasets but also delivers better performance.

Our main contributions are summarized as follows:
\begin{itemize}
    \item We propose a generalized differential equation-driven neural network framework named NvoFDE that incorporates a variable-order differential operator $\frac{\mathrm{d}^{\altalpha(t,\bx)}}{\mathrm{d} t^{\altalpha(t,\bx)}}$. This framework extends the prior classes of constant integer- and fractional-order neural differential equations, subsuming them as special cases with $\altalpha(t,\bx)\equiv\altalpha$ being a constant integer or a constant positive real value, respectively. 
   The order $\altalpha(t,\bx(t))$ is learnable and dependent on both time $t$ and the hidden feature $\bx(t)$.
    Our approach enables flexible and learnable hidden feature updating dynamics stemming from the adaptive differential operator throughout the updating process.
    \item We apply NvoFDE to graph neural networks, extending the capabilities of existing constant-order fractional GNNs. We demonstrate the effectiveness of our framework across multiple graph datasets, showcasing its superior adaptability and performance.
    % \item We apply NvoFDE to variable-order FDE solving where the system variable order is learnable given observations.
    \item Our application of NvoFDE introduces a dynamic approach for solving variable-order FDEs, where the system's order is learnable from empirical observations. This enhancement improves the model's adaptability and predictive accuracy in complex scenarios.
    % NvoFDE introduces a dynamic capability for solving variable-order FDEs, where the system’s order is learnable from empirical observations.
\end{itemize}
    % \item Our empirical evaluations demonstrate the superior performance of NvoFDE in adapting to complex feature dynamics when compared to standard constant order models. This is evidenced by improved accuracy across a range of synthetic and real-world datasets.
    
The structure of this paper is as follows:
We provide a comprehensive overview of related research in the Appendix.
In \cref{sec.prelim}, we review the mathematical foundations of fractional calculus, underscoring the shift from constant to variable orders and its significance for dynamic system modeling.
In \cref{sec.framework}, we describe the architecture of NvoFDE, highlighting its unique features and the reasoning behind its design.
\cref{sec.exp} includes our experimental setup, results, and a detailed analysis of our findings. The implementation code is available at \url{https://github.com/cuiwjTech/AAAI2025_NvoFDE}.
Finally, we conclude the paper and discuss the potential implications of our work for future research in \cref{sec.conclusion}.

% \section{Related Work}
% \label{sec.related}

% \subsection{Fractional Calculus and Its Applications}

% \subsection{Neural Networks with Differential Equations}

\section{Preliminaries}
\label{sec.prelim}
% However, it is usually diﬃcult to obtain the analytical  solution of VO-FDEs. In general, the numerical methods are employed as eﬃcient developed methods for the numerical approximation of VO-FDEs [27, 90]. It is notable to mention that the derivation of an analytical solution for VO-FDE is still in its infancy due to the definitions of VO fractional operators. Hence, many numerical approximation methods and computational techniques have been suggested to investigate the VO-FDE models.

 In this section, we offer a concise overview of fractional calculus. Throughout this paper, we assume that all necessary conditions are met to ensure the well-posedness of the formulations discussed.
 
\noindent\textbf{Cauchy Formula for Integer Order Integration:} We define the operator $J_{t_0}$ as mapping a function $f$, Riemann integrable over $[t_0, t_1]$, to its integral from $t_0$ to $t$: $J_{t_0} f(t)\coloneqq \int_{t_0}^t f(\tau) \mathrm{d} \tau$, for every $t\in[t_0,t_1]$. For $n \in \mathbb{N}$,  the notation $J_{t_0}^n$ represents the $n$-fold iteration of $J_{t_0}$ i.e. we set $J_{t_0}^n\coloneqq J_{t_0} J_{t_0}^{n-1}$ with $J_{t_0}^1\coloneqq J_{t_0}$. According to \cite{diethelm2010analysis}[Lemma 1.1], induction can be employed to show equivalently that:
\begin{align}
    J_{t_0}^n f(t)=\frac{1}{(n-1)!} \int_{t_0}^t(t-\tau)^{n-1} f(\tau) \mathrm{d} \tau,\quad  n \in \mathbb{N}. \label{eq.Cauchy}
\end{align}
\textbf{Riemann-Liouville Fractional Integral Operator:} Let $\altalpha \in \mathbb{R}_{+}$. By extending  $\mathbb{N}$ in \cref{eq.Cauchy} to $\mathbb{R}_{+}$, the Riemann-Liouville fractional integral operator of order $\altalpha$ is naturally defined as
\begin{align}
J_{t_0}^\altalpha f(t) := \frac{1}{\Gamma(\altalpha)} \int_{t_0}^t (t-\tau)^{\altalpha-1} f(\tau) \mathrm{d} \tau, \quad  \altalpha \in \mathbb{R}_{+},\label{eq.RL_int}
\end{align}
where $\Gamma(\cdot)$ is the Gamma function. When $\altalpha=0$, we define $J_{t_0}^0 \coloneqq I$, the identity operator. The generalized variable-order Riemann-Liouville time integration operator whose fractional order varies with time is defined as 
\begin{align}
J_{t_0}^{\altalpha(t)} f(t) \coloneqq \frac{1}{\Gamma(\altalpha(t))} \int_{t_0}^t (t-\tau)^{\altalpha(t)-1} f(\tau) \mathrm{d} \tau, ~ \altalpha(t) \in \mathbb{R}_{+}.\label{eq.RL_int_var}
\end{align}
Here, $\altalpha(t)$ represents the dynamically varying order, allowing for a flexible adjustment of the integration process to accommodate changes in the behavior or characteristics of the function $f(t)$ over time.

\noindent\textbf{(Constant-Order) Caputo Fractional Derivative:} The Caputo fractional derivative is particularly useful in various application fields because it maintains the same initial conditions as traditional integer order differential equations. The Caputo fractional derivative of a function $f(t)$ is defined as:
\begin{align*}
_{t_0}D_t^\altalpha f(t)\coloneqq J_{t_0}^{n-\altalpha} f^{[n]}  =\frac{1}{\Gamma(n-\altalpha)} \int_{t_0}^t \frac{f^{[n]}(\tau)}{(t-\tau)^{\altalpha-n+1}} d \tau, 
\end{align*}
where $n \in \mathbb{N}$ is such that $n-1<\altalpha\leq n$, and $f^{[n]}$ represents the $n$-th integer order derivative $\frac{\ud^n f}{\ud^n t}$. When $\altalpha=n \in \mathbb{N}$, the Caputo fractional derivative simplifies to the classical derivative, i.e., $_{t_0}D_t^\altalpha f(t)\equiv f^{[n]}$.

\noindent\textbf{\emph{Notation Clarification:}} 
% Without loss of generality, our NvoFDE framework, introduced in \cref{sec.framework}, focuses on the interval $[0,T]$, where $T$ represents the end time of integration, rather than an arbitrary interval $[a,b]$. Consequently, we omit the symbol $a$ in the fractional integral or derivative notations, which, however, commonly appear in fractional calculus literature \cite{sun2019review,diethelm2010analysis} since the starting point is also very important for fractional operators. 
In \cref{sec.intro}, we use the fractional derivative notation $\frac{\mathrm{d}^\altalpha}{\mathrm{d} t^\altalpha}$ for readers who may not be familiar with the derivative $_{t_0}D_t^\altalpha$. Moving forward, we will consistently use $_{t_0}D_t^\altalpha$ to denote the fractional derivative to streamline our notation.
\begin{Remark}
From the definition of $_{t_0}D_t^\altalpha f(t)$, it is evident that fractional derivatives incorporate the historical states of the function via the integral term, highlighting their non-local, memory-dependent nature. In contrast, the integer-order derivative only represents the local rate of change of the function.
\end{Remark}

\noindent\textbf{\emph{Variable-Order} Caputo Fractional Derivative:} When the fractional order is allowed to vary with time, the generalized \emph{variable-order} Caputo fractional derivative can be written as below
\begin{align}
_{t_0}D_t^{\altalpha(t)} f(t)\coloneqq \frac{1}{\Gamma(n-\altalpha(t))} \int_{t_0}^t \frac{f^{[n]}(\tau)}{(t-\tau)^{\altalpha(t)-n+1}} d \tau,  \label{eq.vo_Caputo}
\end{align}
with $n-1<\altalpha(t) \leq n$. %Furthermore, $\altalpha(t)$ can be extend to more general $\altalpha(t, y(t))$ with.

Compared to the constant-order Caputo fractional derivative, \cref{eq.vo_Caputo} employs a non-stationary power-law kernel to dynamically adjust its memory structure, influenced by previous values of the differentiation orders. Owing to these properties, the \emph{variable-order} Caputo fractional derivative exhibits a unique memory feature, enabling it to accurately characterize complex physical systems and processes \cite{coimbra2003mechanics,obembe2017variable}. In more general settings, the $\altalpha(t)$ in \cref{eq.RL_int_var,eq.vo_Caputo} can be extended to $\altalpha(t,\cdot)$, wallowing for dependence on parameters beyond $t$.
\noindent\textbf{Constant-Order Fractional Graph Neural Network:}
Recent dynamic process-inspired continuous GNNs \cite{KanZhaDin:C24,ZhaKanSon:C24,chamberlain2021grand,SonKanWan:C22,ZhaKanSon:C23, KanZhaSon:C23,KanLiZha:C25} are based on constant-order FDEs. 
We provide a brief introduction here and will apply our framework to initialize new continuous GNNs based on variable-order FDEs in \cref{ssec.voGNN}.

An undirected graph is denoted as $\calG = (\calV, \bW)$, where $\mathcal{V}$ is the set of $|\calV|$ nodes and $\bY = \left(\left[\by^{(1)}\right]\T, \ldots, \left[\by^{(|\calV|)}\right]\T\right)\T \in \mathbb{R}^{|\calV| \times d}$ consists of the node feature vectors $\by^{(i)} \in \mathbb{R}^{1 \times d}$. The adjacency matrix $\bW$, an $|\calV| \times |\calV|$ matrix, has elements $W_{ij}$ representing the edge weight between the $i$-th and $j$-th nodes with $W_{ij} = W_{ji}$.
In the continuous GNNs, we define $\bY(t) = \left(\left[\by^{(1)}(t)\right]\T, \ldots, \left[\by^{(|\calV|)}(t)\right]\T\right)\T \in \mathbb{R}^{|\calV| \times d}$ as the node features at time $t$, with $\bY(0) = \bY$ as the initial condition. Here, the time $t$ serves as an analog to the layer index \cite{chen2018neural,chamrowgor:grand2021,KanZhaDin:C24}. The dynamics of node features are typically described by the equation:
\begin{align}
{}_0 D_t^{\altalpha} \bY(t) = \calF(\bW,\bY(t)), \text{ with constant } \altalpha, \label{eq.frond_main}
\end{align}
where the function $\calF$ is specifically designed for graph dynamics. For example, in fractional diffusion-inspired GNN models \cite{chamrowgor:grand2021,KanZhaDin:C24}, $\calF$ is defined as $\calF(\bW,\bY(t))=(\mathbf{A}(\mathbf{Y}(t))-\mathbf{I}) \mathbf{Y}(t)$, where $\mathbf{A}(\mathbf{Y}(t))$ is either a learnable attention matrix or a fixed normalized matrix, and $\mathbf{I}$ is an identity matrix. By setting an integration time $T$ and integrating this equation, the updated feature matrix $\bY(t)$ can be obtained, which can subsequently be utilized for downstream tasks such as node classification.

\section{Neural Variable-Order Fractional Differential Equation Networks}
\label{sec.framework}

In this section, we introduce NvoFDE, which utilizes a neural network to parameterize the variable-order fractional derivative of the hidden state. This approach allows for the integration of a continuum of past states into the present state, enabling rich and flexible modeling of hidden features. 
We then outline a numerical scheme to solve this model. 
Furthermore, we present variable-order counterparts of several established constant-order continuous fractional GNNs. 
Additionally, we explore a detailed application of NvoFDE within the PINN methodology \cite{raissi2019physics}, effectively solving variable-order FDEs.

\subsection{Framework}
We denote the hidden feature as $\bx(t)$, where similar to \cite{chen2018neural,KanZhaDin:C24}, the time $t$ serves as an analog to the layer index. 
We propose the following continuous hidden feature updating scheme as a variable-order FDE:
\begin{align}\label{eq.main_fde}
_{t_0}D_t^{\altalpha(t,\bx(t))} \bx(t) = f_{\btheta}(t, \bx(t)), \quad t_0 \leq t \leq t_1,
\end{align}
where $f_{\btheta}(t, \bx(t))$ is a parameterized neural network that outputs the variable-order fractional derivative value at time $t$ for hidden state $\bx(t)$. 
In NvoFDE, the differential order $\altalpha(t,\bx(t))$ is determined by both $t$ and $\bx(t)$, and can be configured as a learnable neural network with scalar output.
This design allows the order to be adaptive, evolving based on time and the dynamics of the hidden features.  
Without loss of generality, we restrict the range of the function $\altalpha(t, \mathbf{x}(t))$ to the interval $(0, 1]$. This is justified by methodologies in \cite{diethelm2010analysis}, which demonstrate that higher-order dynamics can be effectively converted to this range. Moreover, we define the initial state of our system with $\mathbf{x}(t_0) = \mathbf{x}_0$, considering $\mathbf{x}_0$ as the initial feature input.

By setting $\altalpha(t,\bx(t))\equiv 1$, our model simplifies to the first-order neural ODE described in \cite{chen2018neural}. Furthermore, setting it to a constant real value $\altalpha(t,\bx(t))\equiv \altalpha$ aligns it with the constant-order neural FDE discussed in \cite{KanZhaDin:C24}. Thus, NvoFDE effectively generalizes the class of fixed integer- or fractional-order neural differential equation models found in the literature.

Alternatively, motivated by equivalence between the differential form and its integral form \cite{diethelm2010analysis} for constant orders, we can formulate a continuous hidden feature updating scheme as a variable-order fractional integral equation (FIE) given by 
\begin{align}\label{eq.main_fie}
\bx(t) = \bx(t_0) + J_{t_0}^{\altalpha(t,\bx(t))} f_{\btheta}(t, \bx(t)).
\end{align}
It is clear that this framework integrates a continuum of past states into the current state and dynamically adjusts its sensitivity to previous hidden features through the variable order $\altalpha(t,\bx(t))$. 
We note that if $\altalpha(t,\bx(t))$ is constant, then \cref{eq.main_fie} reduces to \cref{eq.main_fde}. To illustrate, if $\altalpha(t,\bx(t))\equiv 1$, the integral equation becomes $\bx(t)=\bx(t_0) + \int_{t_0}^{t}f_{\btheta}(t, \bx(t))$, which is equivalent to the differential equation $\frac{\ud}{\ud t}\bx(t) = f_{\btheta}(t, \bx(t))$. For a general $\altalpha(t,\bx(t))$, the scenarios differ because $\altalpha(t,\bx(t))$ is time-dependent, and transitioning between differential and integral operators is non-trivial.

\subsection{Designing $\altalpha(t,\bx(t))$}
In this section, we explore various designs for the function $\altalpha(t,\bx(t))$. One straightforward approach is to concatenate the time variable $t$ with the state variable $\bx(t)$ and process this combined vector through a Multilayer Perceptron (MLP) or convolutional layers to produce a scalar value representing the differential order. To ensure that $\altalpha(t,\bx(t))$ remains within the interval $(0, 1]$, we can apply a sigmoid activation function at the output. Alternatively, a more sophisticated method involves using the Transformer's sinusoidal position embedding \cite{vaswani2017attention} for the time variable $t$. This embedding can be adjusted to match the dimensionality of $\bx(t)$, allowing for the direct addition of the time position embedding to $\bx(t)$. The enhanced state representation is then processed through either an MLP or convolutional layer to yield a scalar. In simpler scenarios, $\altalpha(t,\bx(t))$ may depend solely on $t$, where $\altalpha(t)$ could be a parameterized function of $t$ or determined by setting learnable values on a uniform grid of $t$ values over the interval $[t_0,t_1]$, with interpolation used to determine $\altalpha(t)$.

\subsection{Solving NvoFDE}\label{ssec.solving NvoFDE}
Existing numerical solvers for integer-order neural ODEs and constant fractional-order neural networks have been extensively studied \cite{chen2018neural,KanZhaDin:C24,diethelm2004detailed}. However, NvoFDE introduces a variable-order FDE/FIE and requires corresponding numerical solutions. In this paper, we examine the explicit L1 solver from \cite{sun2019review} for \cref{eq.main_fde} and two variants of the explicit Adams-Bashforth-Moulton (ABM) solver from \cite{moghaddam2016extended} for \cref{eq.main_fie}. Here, we discuss the L1 predictor as well as the ABM predictor variant. A comprehensive description of the ABM predictor-corrector variant is provided in the Appendix. 

In the following discussion, we consider a uniform grid spanning $[t_0, t_1]$ defined by $\set{t_{n} = t_0 + nh \given n = 0, \dots, N}$, where $h = \frac{t_1 - t_0}{N}$. Furthermore, let $\bx_{n}$ be an approximation of $\bx(t_n)$.

\noindent\textbf{L1 Predictor:} The solution to \cref{eq.main_fde} based on the L1 approximation of the variable-order Caputo fractional derivative \cite{sun2019review} is formulated as:
\begin{align}\label{L1}
\bx_{n+1}=\sum_{j=0}^na_{j,n+1}\bx_j+c_{n+1}f_\theta(t_{n},\bx_{n}),\end{align}
where $c_{n+1}=\Gamma(2-\altalpha(t_n,\bx_n))h^{\altalpha(t_n,\bx_n)}$, 
$a_{0,n+1} = (n+1)^{1-\altalpha(t_n,\bx_n)}-n^{1-\altalpha(t_n,\bx_n)}$, and $a_{j,n+1} = 2(n+1-j)^{1-\altalpha(t_n,\bx_n)}-(n-j)^{1-\altalpha(t_n,\bx_n)}
-(n+2-j)^{1-\altalpha(t_n,\bx_n)}$.
% \begin{align*}
% a_{0,n+1} = &~(n+1)^{1-\altalpha(t_n,\bx_n)}-n^{1-\altalpha(t_n,\bx_n)},\\
% a_{j,n+1} = &~2(n+1-j)^{1-\altalpha(t_n,\bx_n)}-(n-j)^{1-\altalpha(t_n,\bx_n)}\\
% & ~-(n+2-j)^{1-\altalpha(t_n,\bx_n)}, \quad 1\leq j\leq n.
% \end{align*}
\noindent\textbf{ABM Predictor:} The preliminary approximation solution to the variable-order FIE \cref{eq.main_fie} in NvoFDE is formulated as \cite{diethelm2004detailed, moghaddam2016extended}:
\begin{align}\label{predictor}
\bx_{n+1} = \bx_0 + \frac{h^{\altalpha(t_{n},\bx_{n})}}{\Gamma({1+\altalpha(t_{n},\bx_{n}}))} \sum_{j=0}^n b_{j,n+1} f_{\btheta}(t_{j}, \bx_{j}),
\end{align}
where $b_{j,n+1} = (n + 1 - j)^{\altalpha(t_{n},\bx_{n})} -(n - j)^{\altalpha(t_{n},\bx_{n})}, 1\leq j\leq n$. Note that \cref{predictor} can be regarded as the fractional Euler’s method.

\subsection{NvoFDE for Solving Variable-Order FDEs}
Since finding the analytical solutions to general FDEs is difficult or impossible, recently, neural networks have been taken as a universal function to approximate the solutions. Variable-order FDEs describe the variable memory of dynamical systems and do well in representing memory characteristics that change with time or space \cite{lorenzo2002variable, sun2011comparative}. In this section, we consider minimization problems with variable-order FDEs as constraints. It is vital to solve variable-order FDEs by a general approach. To this end, we aim to solve variable-order FDEs using the NvoFDE framework. Existing numerical methods that leverage neural networks to solve variable-order FDEs typically rely on specific mathematical formulations of the order $\altalpha(t, \bx(t))$, often assuming $\altalpha(t, \bx(t))$ to be a linear or periodic function over time \cite{moghaddam2016extended, sun2011comparative}. In contrast, our approach departs from these restrictive assumptions on $\altalpha(t, \bx(t))$, offering a more flexible and generalized treatment.
% In contrast, we explore the unrestricted formulation of $\alpha(t, \bx(t))$, moving beyond its specific mathematical expressions and introducing a novel approach.
% we utilize the truncated power series expansion to replace the unknown functions $u(x)$ in the FDE. 

We first consider the following variable-order FDE:
\begin{align}\label{FDE}
\sum_{k=0}^m Q_k(t) _{t_0}D_t^{\altalpha_k(t)} [u(t)] = h(t,u(t)), \quad t \in [t_0, t_1],
\end{align}
with the initial value $u(t_0)$ and $m$ being a constant. Here, the undetermined \( \altalpha_k(t) \) is the fractional order varying with $t$, while $Q_k(t)$ and $h(t,u)$ are given real-valued analytical functions. Since finding analytical solutions to FDEs is challenging, we focus on using power series expansions to approximate the solution. Thus, we express $u(t)$ as $u(t) = u(0) + \sum_{i=0}^r a_i t^i$, where $r$ is a positive constant that tends to infinity, and $a_i$ are the unknown coefficients to be determined. In practice, we consider the truncated power series expansion of $u(t)$, specifically setting $r$ to 5 in our experiment. When neural networks are used to solve the numerical solution to \eqref{FDE}, the loss function is defined as  
\begin{align*}
\mathcal{J}[u] =  \norm*{\sum_{k=0}^m Q_k(t) _{t_0}D_t^{\altalpha_k(t)}[ u(0) + \sum_{i=0}^r a_i t^i]-h(t,u(t))}, 
\end{align*}
aiming to minimize a functional of $u$, i.e., $\mathcal{J}[u]$, under the constraint \eqref{FDE}. A common algorithm is to solve \eqref{FDE} numerically (e.g., using the L1 or ABM Predictor), and then optimize $\mathcal{J}[u]$ with respect to $\altalpha_k(t)$ and other possible parameters. The above functional $\mathcal{J}[u]$ is equivalent to 
\begin{equation}\label{1.71}
\mathcal{J}[u] =  \norm*{ \sum_{k=0}^{m} Q_k(t)  \zeta_{k}(t) - h(t,u)}^2
\end{equation}
with $\zeta_{k}(t) = \sum_{i=0}^{r} \frac{\Gamma(i + 1)}{\Gamma(i + 1 - \altalpha_k(t))} a_i t^{i - \altalpha_k(t)}$, where $a_i$ are unknown coefficients that will be optimized using the neural networks introduced later. For \eqref{1.71}, we utilize the variable-order fractional derivative expression of each power term $t^n$ for $0 < \altalpha(t) \leq 1$ \cite{akgul2017solutions}:
\begin{align*}
{}_0 D_t^{\altalpha(t)} t^n = \begin{cases}
0, & \text{if } n = 0, \\
\frac{\Gamma(n+1)}{\Gamma(n+1-\altalpha(t))} t^{n-\altalpha(t)}, & \text{if } n \in \mathbb{N},
\end{cases}
\end{align*}
where $\mathbb{N}$ denotes the set of non-negative integers. Therefore, we approximate $\mathcal{J}[u]$ as follows:
\begin{align}\label{1.7}
L_{\text{eqn}} = \sum_{j=1}^N\left( \sum_{k=0}^{m} Q_k(t_j)  \zeta_{k}(t_j) - h(t_j,u(t_j)) \right)^2,
\end{align}
which is referred to as the equation residual for \eqref{FDE}. 
% For a more detailed discussion, we refer the reader to the Appendix. 
% It is also worth noting that differential equations may sometimes involve additional terms such as boundary loss and structure-preserving loss, which should be determined based on the specific physical conditions and inherent properties of the equations \cite{raissi2017physics, chu2024}.

\begin{figure}
      \centering
      \includegraphics[width=0.47\textwidth]{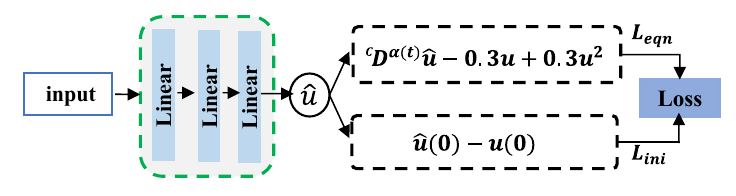}
      \caption{NvoFDE for the Verhulst-Pearl equation. Taking time $t$ as the input to the neural network, $\hat{u}$ is obtained as the output. On the one hand, $\hat{u}$ is involved in \eqref{1.7} and \eqref{VP equation} to compute the equation residual $L_{\text{eqn}}$ by virtue of the ABM predictor; on the other hand, $\hat{u}$ is used to calculate the initial condition residual $L_{\text{ini}}$ based on the initial value of \eqref{VP equation}.}
      \label{fig1}
\end{figure}

As an example, we consider the following variable-order FDE, which is the well-known Verhulst-Pearl equation \cite{moghaddam2016extended}: 
\begin{align}\label{VP equation}
    \begin{cases}
        {_{0}D^{\altalpha(t)}_t} u(t) = 0.3 u(t) - 0.3 u(t)u(t-\delta), & 0 < t \leq T, \\
        u(t) = 0.1, & -\delta \leq t \leq 0,
    \end{cases}
\end{align}
where $\delta$ is a constant delay. The equation \cref{VP equation} models the population growth under environmental constraints. In this paper, we study the case where $\delta = 0$ and solve the equation by NvoFDE. Our framework is presented in \cref{fig1}. 

For \eqref{VP equation}, since the solution $u(t)$ is a function that only involves time $t$, we treat $t$ as the input for the entire neural network. Specifically, we first uniformly select $t_j$ points on the time interval $[0, T]$ as the input with $j$ = 10, 20, 30, 40, etc. For simplicity, we set $T = 1$ in the experiment. The network then gives the output $\hat{u}(t)$. Considering the initial condition $u(0)$ = 0.1 in \eqref{VP equation}, the total loss function of \eqref{VP equation} consists of two components. The first component is the equation residual $L_{\text{eqn}}$ defined in \eqref{1.7}. Based on \cref{VP equation}, it can be seen that $h(t,u)=0.3u-0.3u^2$, $Q_k(t)=1$ and $m=1$ in \eqref{1.7}, where $u(t)$ is provided by the ABM predictor solver discussed in Section \ref{ssec.solving NvoFDE}. The second component is the initial condition residual $L_{\text{ini}}$, calculated as $L_{\text{ini}} = \sum_{i=1}^{N} \big\|\hat{u}_i - u(0)\big\|^2$.
% \begin{align}\label{ini equation}
% L_{\text{ini}} = \sum_{i=1}^{N} \big\|\hat{u}_i - u(0)\big\|^2.
% \end{align}
Thus, we have the total loss function $L_{\text{total}} = \lambda_1 L_{\text{eqn}} + \lambda_2 L_{\text{ini}}$, where $\lambda_1$ and $\lambda_2$ are predefined positive weights.

% The equation residual is given by:
% \begin{align}\label{1.7}
% L_{\text{eqn}} = \norm*{\sum_{k=0}^{m} Q_k(t_j)  \zeta_{k,j} - h(t_j)}^2. 
% \end{align}
% where 
% \begin{align*}
% \zeta_{k,j} = \sum_{i=0}^{r} \frac{\Gamma(i + 1)}{\Gamma(i + 1 - \alpha_k)} a_i t_j^{i - \alpha_k},
% \end{align*}
% the parameter $a_i$ can be trained by the neural network, $Q_k=1$, $h(t) = 0.3 u(t) - 0.3 u^2(t)$, and $u(t)$ can be demonstrated by our basic predictor solver obtained in Section \ref{ssec.solving NvoFDE}. The initial condition residual is calculated:

\subsection{NvoFDE for GNNs} \label{ssec.voGNN}
Consider a graph $\calG = (\calV, \bW)$ as defined in \cref{sec.prelim}. In contrast to the constant-order FROND model in \cite{KanZhaDin:C24}, we introduce a learnable fractional-order function $\altalpha(t,\bY(t))$ to dynamically capture the memory mechanism of feature updating. This new GNN framework is called variable-order FROND (V-FROND). The dynamics of information propagation and feature updating in V-FROND are governed by the following NvoFDE:
\begin{align}\label{1.6}
_{t_0}D_t^{\altalpha(t,\bY(t))} \bY(t) = \calF(\bW,\bY(t)),
\end{align}
where $0 < \altalpha(t,\bY(t)) \leq 1$, and the initial state $\bY(0) = \bY$ consists of the input initial node features. Here, $\calF$ is the dynamic operator on the graph, as described in \cref{sec.prelim}.
By setting an integration time $T$ and integrating this equation, the updated feature matrix $\bY(t)$ can be obtained using \cref{predictor}, which can subsequently be utilized for downstream tasks such as node classification.
\begin{figure*}[!ht]
      \centering
      \includegraphics[width=0.86\textwidth]{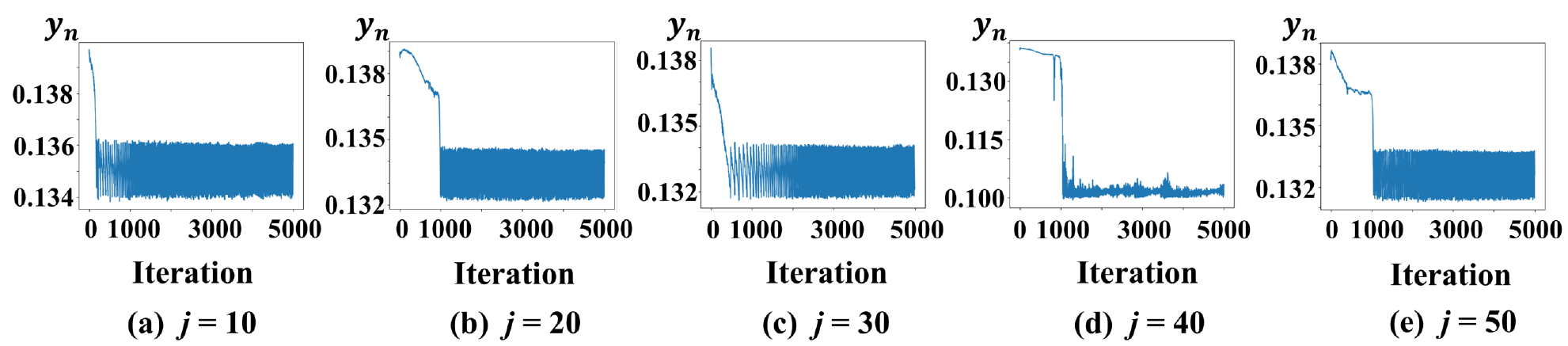}
      \caption{Numerical solutions of the Verhulst-Pearl equation over iterations on the evolution time [0, 1]}
      \label{Numerical solutions}
\end{figure*}

Inspired by the models in \cite{KanZhaDin:C24}, we develop V-FROND variants, including Nvo-GRAND and Nvo-CDE. Similar to \cite{chamberlain2021grand}, Nvo-GRAND includes two versions. One is Nvo-GRAND-nl:
\begin{align}\label{Nvo-GRAND-nl}
_{t_0}D_t^{\altalpha(t,\bY(t))} \bY(t) = (\bA(\bY(t)) - \bI)\bY(t),
\end{align}
where $\bA(\bY(t)) = (a_{i,j}(t))$ is given by a nonlinear attention mechanism as detailed in the Appendix.
% \begin{align*}
% a_{i,j} = \text{softmax} \left( \frac{(\bW_K \by^{(i)}(t)^{\top})^{\top} \bW_Q \by^{(j)}(t)^{\top}}{\sqrt{\bar{d}_k}} \right), 
% \end{align*}
% and $\bW_K$ and $\bW_Q$ are the learned matrices, with $\bar{d}_k$ representing a hyperparameter associated with the dimension of $\bW_K$.
The other version is Nvo-GRAND-l, defined as: 
\begin{align}\label{F-GRAND-l}
_{t_0}D_t^{\altalpha(t,\bY(t))} \bY(t) = -\bL \bY(t), \quad 0 < \altalpha(t,\bY(t)) \leq 1. 
\end{align}
where $\bL$ is a time-invariant matrix, as detailed in the Appendix. This is a linear FDE. %The initial conditions for both of the above versions are set to $\bY(0) = \bY$.

Furthermore, building upon the CDE model \cite{ZhaKanSon:C23}, the Nvo-CDE model is defined as follows:
\begin{align}\label{Nvo-CDE}
_{t_0}D_t^{\altalpha(t,\bY(t))} \bY(t) &= \left(\bA(\bY(t)) - \bI\right)\bY(t)\nonumber\\
&+ \text{div}\left(\bV(t) \circ \bY(t)\right),
\end{align}
where $\text{div}(\cdot)$ is the divergence operator provided by \cite{SonKanWan:C22}, and $\circ$ represents the element-wise (Hadamard) product. This model is crafted to handle heterophilic graphs, where connected nodes belong to different classes or have distinct features. 
In \cref{sec.exp}, we demonstrate the effectiveness of the proposed variable-order FROND framework across multiple graph datasets, showcasing its superior adaptability and performance.

\section{Experiments}
\label{sec.exp}
In this section, we conduct a series of experiments to demonstrate the effectiveness of our proposed methods. All experiments are implemented using the PyTorch framework \cite{paszke2019pytorch} on a single NVIDIA RTX A4000 16GB GPU. For further details, please refer to the Appendix. 
%\vspace{-0.3cm}
\subsection{Experiments on the Verhulst-Pearl equation} \label{ssec.fde_solving}
\textbf{Datasets and Training details}. For the training set, we discretize the time interval $[0, 1]$ uniformly for simplicity, obtaining points $t_j$ where $j$ is a positive integer with $j$ = 10, 20, 30, 40, etc. We use the Adam algorithm \cite{kingma2014adam} with iterations of 200, 500, 1000, 1500 and 2000. The learning rate is set to 0.01. The model employs a three-layer neural network, as shown in Fig.\ref{fig1}, with the hidden layer consisting of 30 neurons. The test set is formed by randomly selecting points $t_j$ from the interval $[0, 1]$, where $j$ takes values like 10, 20, 30, 40, and so on.\\
\textbf{Performance and Analysis}. As seen in Table \ref{loss of VP}, the loss value decreases significantly with the number of iterations. When the step size decreases, which corresponds to an increase of discrete points $j$, it can be observed that better results are obtained for $j = 20$ and $j = 40$. Additionally, during the experiments, it was found that for $\text{Iter} = 2000$, the numerical result reached an accuracy of $10^{-6}$ in 28 instances for $j = 20$ (with results printed every 30 iterations), and in 40 instances for $j = 40$. This indicates that the model is more stable while maintaining high accuracy for $j = 40$. 

In Fig \ref{Numerical solutions}, we present the numerical solution of the Verhulst-Pearl equation over multiple iterations. By varying the step sizes ($j=10, 20, 30, 40, 50$), we observe that after approximately 1000 iterations, all solutions reach a relatively stable state, exhibiting periodic variations. When $j=40$, corresponding to a step size of $1/40$, the amplitude of the solution significantly diminishes after 1000 iterations compared to the solutions with other step sizes. This suggests the numerical solution tends towards a more stable state when $j=40$, which is consistent with the loss analysis discussed above.  

\begin{table}[!htp]
 \fontsize{9pt}{10pt}\selectfont
    \centering
\makebox[0.4\textwidth][c]{
\begin{tabular}{cccccc}
\hline
Iter & $j=10$ & $j=20$ & $j=30$ & $j=40$ & $j=50$ \\
\hline
200 & 8.39E-4 & 5.38E-4 & 4.66E-4 & 5.00E-4 & 4.01E-4 \\
500 & 2.38E-4 & 3.79E-5 & 1.93E-4 & 7.38E-5 & 1.99E-4 \\
1000 & 2.43E-5 & 4.62E-5 & 8.21E-5 & 9.60E-5 & 3.14E-5 \\
1500 & 1.67E-5 & 7.46E-6 & 8.15E-6 & 8.60E-6 & 3.71E-5 \\
2000 & 3.69E-6 & 4.50E-6 & 3.33E-6 & 4.58E-6 & 8.34E-6 \\
\hline
\end{tabular}
}
\caption{Test loss of the Verhulst-Pearl equation over iterations on the evolution time $[0, 1]$}
\label{loss of VP}
\end{table}

% \begin{figure*}[htbp]
% \centering
% % 使用subfigure环境来并排放置图片
% \begin{subfigure}[t]{0.18\textwidth}
%   \includegraphics[width=\textwidth]{Figures/solution y/1.png}
%   \caption{$j$ = 10}
%   \label{fig:1}
% \end{subfigure}
% \hfill % 添加水平填充
% \begin{subfigure}[t]{0.18\textwidth}
%   \includegraphics[width=\textwidth]{Figures/solution y/2.png}
%   \caption{$j$ = 20}
%   \label{fig:2}
% \end{subfigure}
% \hfill
% \begin{subfigure}[t]{0.18\textwidth}
%   \includegraphics[width=\textwidth]{Figures/solution y/3.png}
%   \caption{$j$ = 30}
%   \label{fig:3}
% \end{subfigure}
% \hfill
% \begin{subfigure}[t]{0.18\textwidth}
%   \includegraphics[width=\textwidth]{Figures/solution y/4.png}
%   \caption{$j$ = 40}
%   \label{fig:4}
% \end{subfigure}
% \hfill
% \begin{subfigure}[t]{0.18\textwidth}
%   \includegraphics[width=\textwidth]{Figures/solution y/5.png}
%   \caption{$j$ = 50}
%   \label{fig:5}
% \end{subfigure}
% \caption{Numerical solutions of the Verhulst-Pearl equation over iterations on the evolution time $[0, 1]$}
% \label{Numerical solutions}
% \end{figure*}

\begin{table*}[!htp]
\centering
% \tiny
\fontsize{9pt}{10pt}\selectfont
\setlength{\tabcolsep}{2pt}
\makebox[\textwidth][c]{
    \begin{tabular}{l|cccccccc|cc}
        \toprule
        Method & Cora & Citeseer & Pubmed & CoauthorCS & Computer & Photo & CoauthorPhy & ogbn-arxiv & Airport & Disease \\
        \midrule
        GCN & 81.5$\pm$1.3 & 71.9$\pm$1.9 & 77.8$\pm$2.9 & 91.1$\pm$0.5 & 82.6$\pm$2.4 & 91.2$\pm$1.2 & 92.8$\pm$1.0 & 72.2$\pm$0.3 & 81.6$\pm$0.6 & 69.8$\pm$0.5\\
        GAT & 81.8$\pm$1.3 & 71.4$\pm$1.9 & 78.7$\pm$2.3 & 90.5$\pm$0.6 & 78.0$\pm$19.0 & 85.7$\pm$20.3 & 92.5$\pm$0.9 & \first{73.7$\pm$0.1} & 81.6$\pm$0.4 & 70.4$\pm$0.5\\
        HGCN & 78.7$\pm$1.0 & 65.8$\pm$2.0 & 76.4$\pm$0.8 & 90.6$\pm$0.3 & 80.6$\pm$1.8 & 88.2$\pm$1.4 & 90.8$\pm$1.5 & 59.6$\pm$0.4 & 85.4$\pm$0.7 & 89.9$\pm$1.1\\
        GIL & 82.1$\pm$1.1 & 71.1$\pm$1.2 & 77.8$\pm$0.6 & 89.4$\pm$1.5 & -- & 89.6$\pm$1.3 & -- & -- & 91.5$\pm$1.7 & 90.8$\pm$0.5\\
        \midrule
        GRAND-l & {83.6$\pm$1.0} & 73.4$\pm$0.5 & 78.8$\pm$1.7 & 92.9$\pm$0.4 & 83.7$\pm$1.2 & 92.3$\pm$0.9 & 93.5$\pm$0.9 & 71.9$\pm$0.2 & 80.5$\pm$9.6 & 74.5$\pm$3.4\\
        F-GRAND-l & {84.8$\pm$1.1} & {74.0$\pm$1.5} & {79.4$\pm$1.5} & \second{93.0$\pm$0.3} & {84.4$\pm$1.5} & {92.8$\pm$0.6} & {94.5$\pm$0.4} & \second{72.6$\pm$0.1} & {98.1$\pm$0.2} &{92.4$\pm$3.9}\\
        Nvo-GRAND-l (ours) & \first{86.0\(\pm\)0.5} & \second{75.6\(\pm\)0.8} & \first{80.8\(\pm\)1.2} & \first{93.4\(\pm\)0.2}& \first{87.9\(\pm\)0.8} & \first{94.1\(\pm\)0.2} & \first{94.7\(\pm\)0.2} & {71.8\(\pm\)0.1} & \first{98.7\(\pm\)0.2} & \first{97.4\(\pm\)0.7}\\
        \midrule
        GRAND-nl & 82.3$\pm$1.6 & 70.9$\pm$1.0 & 77.5$\pm$1.8 & 92.4$\pm$0.3 & 82.4$\pm$2.1 & 92.4$\pm$0.8 & 91.4$\pm$1.3 & 71.2$\pm$0.2 & 90.9$\pm$1.6 & 81.0$\pm$6.7\\
        F-GRAND-nl & {83.2$\pm$1.1} & {74.7$\pm$1.9} & {79.2$\pm$0.7} & {92.9$\pm$0.4} & {84.1$\pm$0.9} & {93.1$\pm$0.9} & {93.9$\pm$0.5} & {71.4$\pm$0.3} & {96.1$\pm$0.7} & {85.5$\pm$2.5}\\
        Nvo-GRAND-nl (ours) & \second{85.4\(\pm\)1.0} & \first{75.9\(\pm\)0.6} & \second{80.6\(\pm\)0.7} & \first{93.4\(\pm\)0.2} & \second{87.2\(\pm\)1.4} & \second{94.0\(\pm\)0.3} & \second{94.6\(\pm\)0.2} & {72.0\(\pm\)0.2} & \second{98.4\(\pm\)0.2} & \second{89.8\(\pm\)3.4} \\
        \bottomrule
    \end{tabular}
    }
    % %\vspace{-0.1cm}
\caption{Node classification results (\%) for random train-val-test splits. The best and the second-best results for each criterion are highlighted in bold and underlined, respectively.}
% %\vspace{-0.2cm}
\label{tab:node_res_homo}
\end{table*}

\begin{table*}[!htp]
\centering
% \tiny
\fontsize{9pt}{10pt}\selectfont
\setlength{\tabcolsep}{4pt}
\makebox[0.8\textwidth][c]{
\begin{tabular}{lcccccc}
\hline
Model & Roman-empire & Wiki-cooc & Minesweeper & Questions & Workers & Amazon-ratings \\
\hline
% GRAND-LAP & 69.24$\pm$0.53 & 91.58$\pm$0.37 & 73.25$\pm$0.99 & 68.54$\pm$1.07 & 75.59$\pm$0.86 & 48.99$\pm$0.35 \\
% GRAND-nl & 71.60$\pm$0.58 & 92.03$\pm$0.46 & 76.67$\pm$0.98 & 70.67$\pm$1.28 & 75.33$\pm$0.84 & 45.05$\pm$0.65 \\
% GRAND-TRANS & 71.18$\pm$0.56 & 91.86$\pm$0.27 & 75.40$\pm$1.36 & 69.14$\pm$0.97 & 75.13$\pm$0.65 & 45.20$\pm$0.52 \\
GRAND-l & 69.24$\pm$0.53 & 91.58$\pm$0.37 & 73.25$\pm$0.99 & 68.54$\pm$1.07 & 75.59$\pm$0.86 & 48.99$\pm$0.35 \\
GRAND-nl & 71.60$\pm$0.58 & 92.03$\pm$0.46 & 76.67$\pm$0.98 & 70.67$\pm$1.28 & 75.33$\pm$0.84 & 45.05$\pm$0.65 \\
GraphBel & 69.47$\pm$0.37 & 90.30$\pm$0.50 & 76.51$\pm$1.03 & 70.79$\pm$0.99 & 73.02$\pm$0.92 & 43.63$\pm$0.42 \\
NSD & 77.50$\pm$0.67 & 92.06$\pm$0.40 & 89.59$\pm$0.61 & 69.25$\pm$1.15 & 79.81$\pm$0.99 & 37.96$\pm$0.20 \\
\hline
% % CDE-GRAND-LAP & 90.58$\pm$0.49 & 98.00$\pm$0.20 & 90.06$\pm$0.60 & 73.78$\pm$1.46 & 80.45$\pm$0.97 & 47.43$\pm$0.53 \\
% CDE-GRAND-nl & 91.64$\pm$0.28 & 97.99$\pm$0.38 & 95.50$\pm$5.23 & \first{75.17$\pm$0.99} & 80.70$\pm$1.04 & 47.63$\pm$0.43 \\
% % CDE-GRAND-TRANS & 91.55$\pm$0.23 & 98.04$\pm$0.35 & 91.38$\pm$1.92 & 72.92$\pm$1.54 & 81.58$\pm$0.98 & 46.91$\pm$0.87 \\
% CDE-GraphBel & 85.39$\pm$0.46 & 97.79$\pm$0.40 & 90.79$\pm$0.48 & 72.11$\pm$1.31 & 81.30$\pm$0.43 & 45.22$\pm$0.60 \\
% CDE-Diag-NSD & 78.99$\pm$0.52 & 92.45$\pm$0.67 & 91.13$\pm$0.80 & 73.65$\pm$1.55 & 82.81$\pm$0.51 & 40.92$\pm$1.96 \\
CDE & 91.64$\pm$0.28 & 97.99$\pm$0.38 & 95.50$\pm$5.23 & \first{75.17$\pm$0.99} & 80.70$\pm$1.04 & 47.63$\pm$0.43 \\
F-CDE & \second{93.06$\pm$0.55} & \second{98.73$\pm$0.68} & \second{96.04$\pm$0.25} & \first{75.17$\pm$0.99} & \second{82.68$\pm$0.86} & \second{49.01$\pm$0.56} \\
Nvo-CDE (ours) & \first{93.42$\pm$0.22} & \first{99.32$\pm$0.28} & \first{98.53$\pm$0.27} & \second{74.87$\pm$0.23} & \first{83.33$\pm$0.65} & \first{50.09$\pm$0.40} \\
\hline
\end{tabular}}
\caption{Node classification results (\%) on large heterophilic datasets.}
%\vspace{-0.2cm}
\label{tab:large_heterophilic}
\end{table*}

\subsection{Node Classification on Homophilic Graph 
}
\begin{figure}
      \centering
      \includegraphics[width=0.43\textwidth]{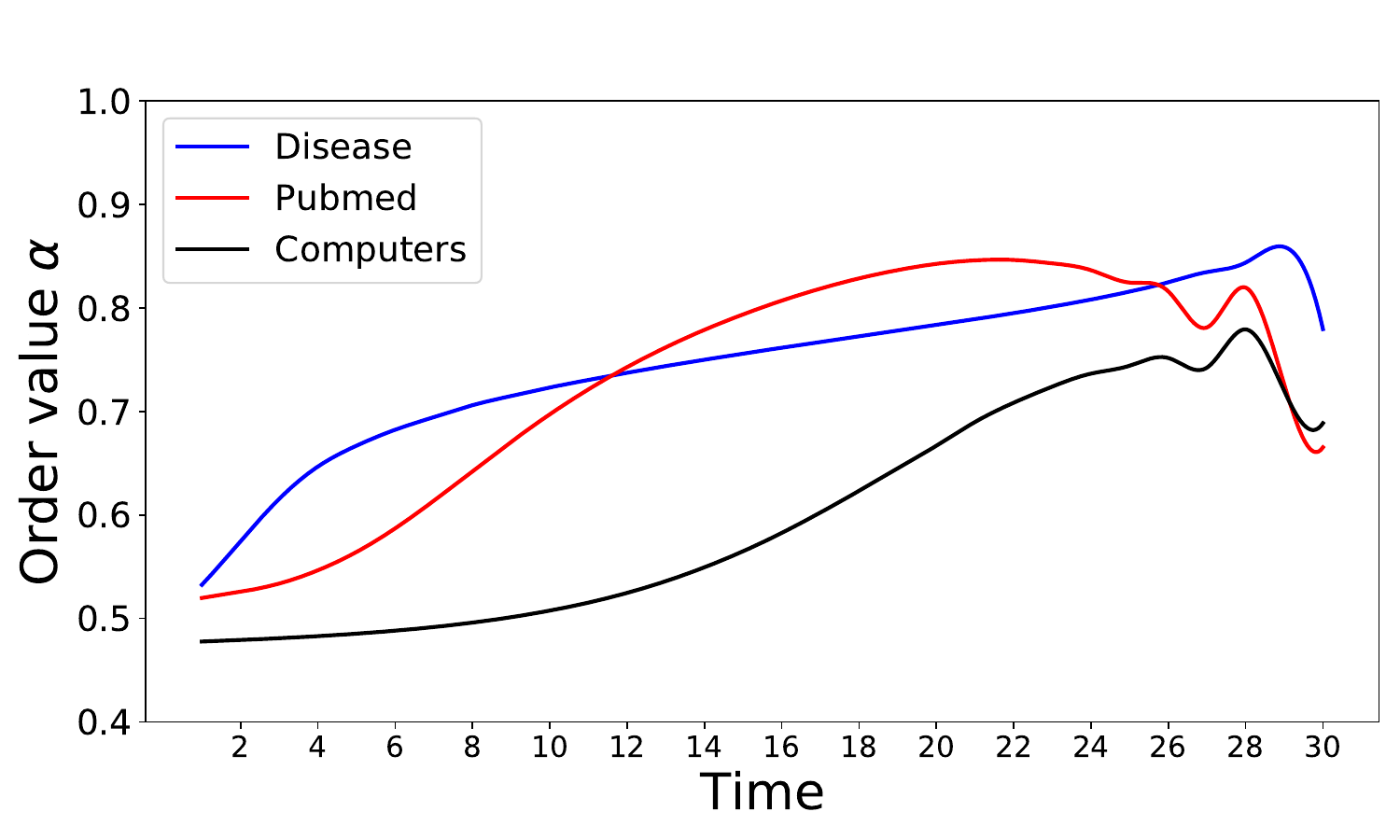}
      \caption{The order value evolution of  $\altalpha(t,\bx(t))$.}
      \label{GNN order value}
\end{figure}

\textbf{Datasets}. Our study encompasses diverse datasets with various topologies. For the Disease and Airport datasets, we employ the same data splitting and pre-processing methods as detailed in \cite{chami2019hyperbolic_GCNN}. For the remaining datasets, we follow the experimental settings used in GRAND \cite{chamrowgor:grand2021} and F-GRAND, applying random splits to the largest connected component of each dataset. For more details regarding the dataset, please refer to the Appendix.\\
\textbf{Methods}. For our comparative analysis, we select several prominent GNN models, including GCN \cite{kipf2017semi}, GAT \cite{velickovic2018graph}, HGCN \cite{chami2019hyperbolic_GCNN}, and GIL \cite{zhu2020GIL}. GRAND \cite{chamrowgor:grand2021} serves as a specific instance of Nvo-GRAND, characterized by the order $\altalpha(t,\bx(t)) = 1$. We examine two variants of Nvo-GRAND: Nvo-GRAND-nl \cref{Nvo-GRAND-nl} and Nvo-GRAND-l \cref{F-GRAND-l}. For consistency in comparison, we report baseline results from the FROND paper \cite{KanZhaDin:C24}. \\
\textbf{Performance and Analysis}.  \cref{tab:node_res_homo} presents the experimental results. As expected, even though F-GRAND already achieves excellent performance, Nvo-GRAND still surpasses F-GRAND on almost all datasets. The main reason lies in Nvo-GRAND’s learnable ability to flexibly adjust its memory mechanism, allowing the model to find local optimal solutions and optimize its memory trajectory. Additionally, we illustrate the evolutionary trend of the order $\altalpha(t,\bx(t))$ in Figure \ref{GNN order value}. We select three different types of datasets, namely Computers, Pubmed and Disease. The initial order is set to $\altalpha(t) \equiv 0.8$ at each time point. After training, we observe that $\altalpha(t,\bx(t))$ has dramatically altered its evolutionary path over time. Our method allows $\altalpha(t,\bx(t))$ to adapt actively, rather than relying on conventional mathematical designs such as linear functions or periodic functions with respect to $t$ as used in prior work \cite{moghaddam2016extended, sun2011comparative}.

\subsection{Node Classification on Heterophilic Graph}
\textbf{Datasets}. The paper \cite{platonov2023critical} demonstrated significant flaws in the standardized datasets used for evaluating models on heterophilic graphs and introduced six heterophilic graph datasets, namely Roman-empire, Wiki-cooc, Minesweeper, Questions, Workers, and Amazon-ratings. These datasets come from various domains, have low homophily scores, and exhibit diverse structural properties. The dataset splits used in this section follow the same approach as in \cite{platonov2023critical}. Consistent with \cite{ZhaKanSon:C23}, the ROC-AUC score is utilized as the evaluation metric for the Minesweeper, Workers, and Questions datasets, given that they entail binary classification. \\
\textbf{Methods}. To ensure a fair comparison, we adhere to the experimental settings outlined in the CDE paper \cite{ZhaKanSon:C23}. We evaluate the performance of the Nvo-CDE model by comparing it against other graph neural ODE models, including GraphBel \cite{SonKanWan:C22}, NSD \cite{crifraben:sheaf2022}, CDE \cite{ZhaKanSon:C23} and F-CDE \cite{KanZhaDin:C24}, as detailed in \cref{tab:large_heterophilic}.\\
\textbf{Performance and Analysis}. From \cref{tab:large_heterophilic}, it is evident that our Nvo-CDE model outperforms competitors on five out of six datasets, demonstrating the efficacy of the Nvo-CDE model.
On the Minesweeper dataset, the use of variable-order derivatives improves performance by 2.49\%. This advantage stems from the framework's flexibility in modeling adaptive derivative orders based on hidden features, capturing more complex feature-updating dynamics.

\subsection{Image Classification of NvoFDE}

In this part, we consider applying the NvoFDE framework to image classification tasks. 
We refer to the adapted model as the NvoFDE-based Image Classification, abbreviated as Nvo-IC. The architecture of Nvo-IC mirrors that of Neural ODEs \cite{chen2018neural}, with a significant modification: the core component is a variable-order FDE block instead of the traditional integer-order ODE block. We evaluate the Nvo-IC model by conducting experiments on the Fashion-MNIST \cite{xiao2017fashion} and CIFAR \cite{KriTR2009} datasets, assessing classification accuracy on both clean and stochastically noisy images. The experimental outcomes indicate that Nvo-IC marginally outperforms traditional models in these tasks, underscoring the efficacy and potential of the NvoFDE framework for complex image classification scenarios. For detailed experimental results and further analysis, please refer to the Appendix.

\section{Conclusion}\label{sec.conclusion}

We present the Neural Variable-Order Fractional Differential Equation network (NvoFDE), a novel framework that integrates variable-order fractional derivatives. 
Our approach enables the modeling of adaptive derivative orders based on hidden features, capturing complex feature-updating dynamics with greater flexibility. 
We apply NvoFDE to graph neural networks, enhancing the capabilities of existing constant-order fractional GNNs. 
Through extensive experiments on multiple graph datasets, we demonstrate the superior adaptability and performance of NvoFDE. Additionally, NvoFDE introduces a dynamic capability for solving variable-order FDEs, where the system’s order is learnable from empirical observations. 
Our framework holds significant potential for various applications, such as learnable fractional viscoelastic modeling and adaptive control with learned orders.

\section*{Acknowledgments}
This work is supported by the National Natural Science Foundation of China under Grant Nos. U2268203, 12301491, 12225107 and 12071195, the Major Science and Technology Projects in Gansu Province-Leading Talents in Science and Technology under Grant No. 23ZDKA0005. This research is also supported by the National Research Foundation, Singapore and Infocomm Media Development Authority under its Future Communications Research and Development Programme.% This work is supported by the National Natural Science Foundation of China under Grant Nos. U2268203, 
% This work is supported by the National Natural Science Foundation of China under Grant Nos. 12301491, 12225107 and 12071195.
% National Natural Science Foundation of China under Grant Nos. 12225107 and 12071195, the Major Science and Technology Projects in Gansu Province-Leading Talents in Science and Technology under Grant No. 23ZDKA0005, the Innovative Groups of Basic Research in Gansu Province under Grant No. 22JR5RA391, and Lanzhou Talent Work Special Fund.
To improve the readability, parts of this paper have been grammatically revised using ChatGPT \cite{openai2022chatgpt4}.

\bibliography{main}
% \bibliographystyle{aaai25}

% \bibliography{references}

%\begin{figure*}
      %\centering
      %\includegraphics[width=0.99\textwidth]{Figures/fulu.pdf}
     % \label{oversmooth}
%\end{figure*}



\section*{Supplementary material}

# Appendix for Neural Variable-Order Fractional Differential Equation Networks

Wenjun Cui, Qiyu Kang, Xuhao Li, Kai Zhao, Wee Peng Tay, Weihua Deng, Yidong Li

This supplementary material complements the main text of our paper, offering detailed information and further evidence for the claims presented. The organization of this document is as follows:

1. A comprehensive review of relevant literature is expanded upon in Section S1.
2. The specifics of the solvers employed in our paper are elaborated in Section S2.
3. We provide additional experimental settings and specifications of the datasets in Section S3.
4. Lastly, we address the limitations of our work and discuss its broader impact in Section S4.

Labels throughout this supplementary material are prefixed with "S" for clarity.

## S1   Related Work

In this section, we provide a review of fractional calculus and its broad applications. Additionally, we offer a detailed overview of the integration of neural networks with fractional differential equations (FDEs).

### S1.1   Fractional Calculus and Its Applications

Fractional calculus extends the concept of derivatives and integrals to non-integer orders, establishing itself as a vital mathematical tool for characterizing the memory properties of complex systems and certain materials. Unlike traditional calculus, which only deals with instantaneous changes, fractional calculus considers the influence of past states on the current state of a system. There are several definitions of fractional derivatives. For example, the most commonly used definitions are the Riemann-Liouville (Munkhammar 2004) and the Caputo-Liouville (Luchko 1999) fractional derivatives, both of which utilize a power-law kernel to weight the past states. Recognizing the limitations of the power law in modeling all physical problems, Caputo and Fabrizio proposed an alternative differential concept that uses exponential decay as a kernel, offering a strong ability to describe material heterogeneity and structures at different scales (Caputo and Fabrizio 2015). Furthermore, Atangana and Baleanu introduced a new fractional derivative based on the Mittag-Leffler function to bridge limitations in the power-law applications (Atangana and Baleanu 2016). Applications of fractional calculus in various domains are extensive. (Mainardi 2022) discussed its application in viscoelasticity, demonstrating that fractional models accurately describe the memory and hereditary properties of such materials. (Podlubny 1999) provided an introduction to the practical applications of fractional calculus in signal processing, modeling, and automatic control. Moreover, fractional calculus has been applied to enhance densely connected graph neural networks, addressing issues like oversmoothing and the vanishing gradient problem (Antil et al. 2020; Kang et al. 2024a).

Further extending the capabilities of constant-order counterparts, variable-order FDEs offer improved descriptions of nonlocal phenomena with varying dynamics. In these systems, the fractional order $\alpha$ varies with time, represented as $\alpha(t)$, allowing for a more accurate modeling of dynamic processes. The concept of variable-order fractional derivatives, first introduced by (Samko and Ross 1993; Coimbra 2003), has proven particularly effective in describing complex memory properties. These derivatives have been utilized in numerous practical applications, such as modeling electrochemical systems, drug reaction processes, and rock creep behavior, effectively capturing the variable dynamics inherent in these phenomena (Sheng et al. 2011; Ramirez and Coimbra 2011; Tang et al. 2018).

### S1.2   Neural Networks with Fractional Differential Equations

Neural networks can be utilized as a universal function for numerically solving FDEs, which is crucial since finding analytical solutions to general FDEs is often challenging. (Jafarian, Mokhtarpour, and Baleanu 2017) proposes an algorithm to approximate the solution of a class of fractional-order initial value ordinary differential equations using neural networks. This approach converts FDEs into a minimization problem by employing a truncated power series of the solution function. (Zúñiga-Aguilar et al. 2018) consider variable-order FDEs in the Liouville-Caputo sense and use the Levenberg-Marquardt algorithm to optimize synaptic weights, obtaining an approximate solution to the equation. The effectiveness of the proposed neural network is demonstrated by comparing it with the analytical solution. Leveraging the advantages of Lagrange polynomials in FDEs, a functional link neural net-

work based on Lagrange polynomials is designed, offering a new approach for solving a class of nonlinear variable-order FDEs (Lasaki, Ebrahimi, and Ilie 2023).

On the other hand, FDEs have also enriched neural networks. Research combining variable-order FDEs with neural networks has been particularly relevant in computational neuroscience (Anastasio 1994) and Hopfield networks (Kaslik and Sivasundaram 2012), focusing on numerical simulations and analyzing bifurcation and stability behavior within connected networks (Rakkiyappan et al. 2017; Yang et al. 2018; Xu et al. 2019). More recently, there has been a deep exploration of the intersection between FDEs and graph neural networks. (Kang et al. 2024a,b) introduces the fractional-order graph neural dynamical network, utilizing the Caputo fractional derivative. This approach enables the capture of long-term dependencies in feature updates, enhancing the capacity for graph representation learning.

## S2  Numerical Solver for NvoFDE

In our work, we employ several numerical solvers for NvoFDE (Sun et al. 2019; Moghaddam, Yaghoobi, and Tenreiro Machado 2016; Diethelm 2010). Section 3.3 elaborates on the classic variable-order L1 Predictor and ABM Predictor used in our experiments. Here, we introduce a variable-order Predictor-Corrector solver, which can enhance the numerical accuracy of the variable-order FDE solutions.

**Predictor-Corrector solver:** We consider a uniform grid. Let $y_h(t_n)$ denote the approximation to $y(t_n)$. Assume that we have already calculated approximations $y_h(t_j)$, $j = 0, 1, 2, \cdots, n$, and want to obtain $y_h(t_{n+1})$ by means of the equation:

$$y_h(t_{n+1}) = y_0 + \frac{h^{\alpha(t_{n+1}, \mathbf{x}_{n+1})}}{\Gamma(\alpha(t_{n+1}, \mathbf{x}_{n+1}) + 2)} f(t_{n+1}, y_h^P(t_{n+1})) + \frac{h^{\alpha(t_{n+1}, \mathbf{x}_{n+1})}}{\Gamma(\alpha(t_{n+1}, \mathbf{x}_{n+1}) + 2)} \sum_{j=0}^{n} a_{j,n+1} f(t_j, y_n(t_j)).$$

where

$$a_{j,n+1} = \begin{cases} n^{\alpha(t_{n+1}, \mathbf{x}_{n+1})+1} \\ -(n - \alpha(t_{n+1}, \mathbf{x}_{n+1}))(n+1)^{\alpha(t_{n+1}, \mathbf{x}_{n+1})}, \\ \quad \text{for} \quad j = 0, \\ (n - j + 2)^{\alpha(t_{n+1}, \mathbf{x}_{n+1})+1} + (n - j)^{\alpha(t_{n+1}, \mathbf{x}_{n+1})+1} \\ -2(n - j + 1)^{\alpha(t_{n+1}, \mathbf{x}_{n+1})+1}, \quad \text{for} \quad 1 \leq j \leq n, \\ 1, \quad \text{for} \quad j = n + 1. \end{cases}$$

The preliminary approximation is called predictor and is given by

$$y_h^P(t_{n+1}) = y_0 + \frac{1}{\Gamma(\alpha(t_{n+1}, \mathbf{x}_{n+1}))} \sum_{j=0}^{n} b_{j,n+1} f(t_j, y_n(t_j)),$$

where

$$b_{j,n+1} = \frac{h^{\alpha(t_{n+1}, \mathbf{x}_{n+1})}}{\alpha(t_{n+1}, \mathbf{x}_{n+1})} \left( (n+1-j)^{\alpha(t_{n+1}, \mathbf{x}_{n+1})} - (n-j)^{\alpha(t_{n+1}, \mathbf{x}_{n+1})} \right)$$

## S3  Experiments

### S3.1  Dataset Description for Solving Variable-Order FDEs

This section details the dataset preparation procedures outlined in Section 4.1, which underpin the experiments discussed. For the training set, we discretized the time interval $[0, 1]$ uniformly, obtaining points $t_j$ where $j$ is a positive integer with $j = 10, 20, 30, 40$, etc. For instance, when $j = 10$, the values of $t_j$ are $0, \frac{1}{9}, \frac{2}{9}, \ldots, 1$. For the test set, the time point $t_j$ was randomly generated within the interval $[0, 1]$, with the number of points also being 10, 20, 30, 40, and so on. In one of our experiments, when the discretization point was set to 10, the randomly generated time points $t_j$ was 0.058, 0.106, 0.156, 0.374, 0.598, 0.601, 0.708, 0.732, 0.866, and 0.951, respectively.

### S3.2  Graph Dataset statistics

Our study utilizes a diverse collection of graph datasets. The statistics for these datasets, as used in Table 2, are detailed in Table S1. Consistent with the experimental framework established in (Chamberlain et al. 2021), we select the largest connected component from each dataset, except for the tree-like graph datasets (Airport and Disease).

| Dataset | Type | Classes | Features | Nodes | Edges |
|---|---|---|---|---|---|
| Cora | citation | 7 | 1433 | 2485 | 5069 |
| Citeseer | citation | 6 | 3703 | 2120 | 3679 |
| PubMed | citation | 3 | 500 | 19717 | 44324 |
| Coauthor CS | co-author | 15 | 6805 | 18333 | 81894 |
| Computers | co-purchase | 10 | 767 | 13381 | 245778 |
| Photos | co-purchase | 8 | 745 | 7487 | 119043 |
| CoauthorPhy | co-author | 5 | 8415 | 34493 | 247962 |
| OGB-Arxiv | citation | 40 | 128 | 169343 | 1166243 |
| Airport | tree-like | 4 | 4 | 3188 | 3188 |
| Disease | tree-like | 2 | 1000 | 1044 | 1043 |

Table S1: Dataset Statistics used in Table 2

### S3.3  Oversmoothing of NvoFDE

We tested the performance of Nvo-GRAND in mitigating the over-smoothing problem by increasing the network depth. Using the ABM Predictor from equation (9), we constructed neural networks with varying depths. We followed the fixed data split method from (Chami et al. 2019; Kang et al. 2024a). The performance was assessed on three datasets: Cora, Citeseer, and Disease. As shown in Fig. S1, our model consistently outperformed both GRAND-l and F-GRAND-l across all datasets. Notably, on the Disease dataset, Nvo-GRAND achieved performance improvements of 6%, 6.5%, 9%, 11%, 7%, and 5% at network depths of 4, 8, 16, 32, 64, and 128 layers, respectively. This enhancement is consistent with the model's flexible memory mechanism, which allows it to autonomously optimize the order $\alpha(t, \mathbf{x}(t))$.

### S3.4  Image Classification of NvoFDE

This section investigates the application of NvoFDE to image classification tasks, with the improved model referred to as

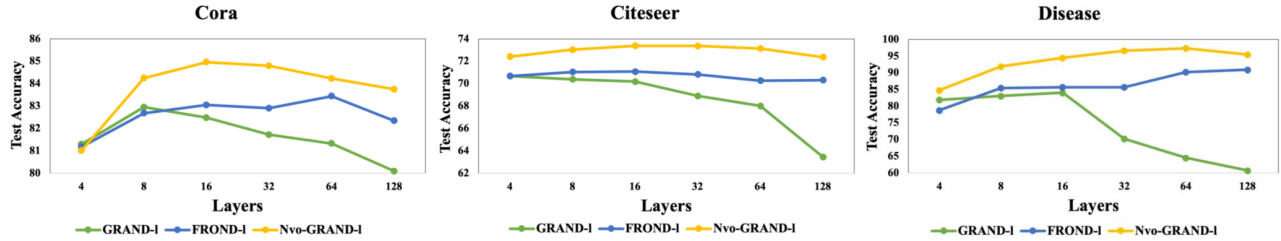

Figure S1: Oversmoothing Mitigation

NvoFDE-based Image Classification (Nvo-IC). While maintaining structural similarities with Neural ODEs (Chen et al. 2018), Nvo-IC replaces the conventional ODE numerical solver with a variable-order FDE numerical solver (detailed in Section 3.3). As a result, Nvo-IC not only inherits the advantages of Neural ODEs but also incorporates a dynamic adaptive adjustment strategy through the variable-order fractional operator, offering a novel perspective for image classification tasks.

We mainly use two benchmark models for comparison. One is Neural ODEs (NODEs), implemented with four numerical solvers: Euler, Midpoint, Bosh3Solver and RK4. The other benchmark is a constant-order FDE-based classification model, referred to as Neural FDE-based Image Classification (NFDE-IC). NFDE-IC serves as the constant-order counterpart of Nvo-IC. Accordingly, the experimental comparison includes five configurations: NODE with Euler, NODE with Midpoint, NODE with Bosh3Solver, NODE with RK4, and NFDEs-IC. Euler, Midpoint, Bosh3Solver, and RK4 have already been presented in (Chen et al. 2018), thus a brief overview of them is provided in the following Remark 1.

**Remark 1.** *Euler's Method: This is the simplest numerical method for solving ODEs. It approximates the solution by taking a straight-line step from the current point using the slope provided by the derivative. Midpoint Method: Also known as the modified Euler method, this algorithm improves the Euler's method by taking an additional intermediate step to better estimate the slope. Bosh3 Solver: It is a third-order method used for solving ODEs. It is more accurate than lower-order methods like Euler and Midpoint. RK4 (Runge-Kutta 4th Order): RK4 is a widely used and highly accurate method for solving ODEs. RK4 is popular for its balance between accuracy and computational cost, making it a first method for many practical problems.*

**Datasets** We conducted evaluations of our model on two distinct classification datasets: Fashion-MNIST (Xiao, Rasul, and Vollgraf 2017) and CIFAR-10 (Krizhevsky and Hinton 2009) benchmarks. The Fashion-MNIST dataset consists of 70,000 grayscale images, each of 28 × 28 pixels, categorized into 10 classes. Out of these, 60,000 images are used for training, and the remaining 10,000 are reserved for testing. In line with (Chen et al. 2018), we applied a straightforward data augmentation approach, which includes adding a 4-pixel padding around each image followed by random cropping. The CIFAR-10 comprises 60,000 color images of 32 × 32 pixels, which is split into 10 separate classes. The dataset is divided into 50,000 training images and 10,000 testing images. We enhanced the dataset using data augmentation techniques outlined in (Yang et al. 2020). These included padding with 4 pixels around each image, image normalization, random horizontal flipping, color jitter and random cropping.

**Training Details** The backward propagation is computed using automatic differentiation in PyTorch. In all experiments, we utilized the SGD optimizer, with a weight decay 1e-4 and momentum fixed at 0.9. The batch size was set to 128. The training was carried out using the CosineAnnealingLR method (Loshchilov and Hutter 2016) with 160 epochs. We tested the classification accuracy of the model on clean images and stochastic noise images. We used four types of noise (specifically, impulse noise, speckle noise, Gaussian noise, and shot noise) from the CIFAR-10-C dataset (Hendrycks and Dietterich 2019) in our experiments. Furthermore, the classification accuracy on stochastic noise images was also evaluated over the Fashion-MNIST dataset.

**Performance and Analysis** As depicted in Table S2, experiments indicate that our model achieves good performance in classification accuracy on both stochastic noise and clean testing images, strongly demonstrating its effectiveness and broad applicability. For instance, on the CIFAR-10 dataset, the average accuracy of our model over the four stochastic noise images outperforms the five baseline models by 4.35%, 0.57%, 1.34%, 1.92% and 0.53%, respectively. By conducting classification tests on stochastic noise images, we can tentatively conclude that our model also possesses a notable degree of robustness.

## S4 Limitations and Broader Impact

While our NvoFDE introduces a significant advancement in modeling feature updating dynamics with variable-order fractional derivatives, several limitations should be acknowledged. For instance, while NvoFDE has shown promising performance on the graph and image datasets tested, the generalization of this model to other types of data or tasks, such as time-series prediction, has not been thoroughly evaluated. This performance variability may necessitate extensive testing across different domains. The broader impact of the NvoFDE model spans both scientific research and practical applications. By enabling the learning and adaptive changes

| Benchmark | Model | Acc. | Impulse | Speckle | Gaussian | Shot | Avg. |
|---|---|---|---|---|---|---|---|
| Fashion-MNIST | NODEs with Euler | 93.62 | 82.82 | 75.51 | 82.73 | 81.81 | 80.72 |
| | NODEs with Midpoint | 93.70 | 83.79 | 75.27 | 83.98 | 82.83 | 81.47 |
| | NODEs with Bosh3Solver | 94.00 | 84.33 | 76.76 | 82.50 | 83.25 | 81.71 |
| | NODEs with RK4 | 94.07 | 84.46 | 77.51 | 82.34 | 83.20 | 81.88 |
| | NFDE-IC | 94.02 | 85.29 | 76.68 | 83.46 | 84.17 | 82.40 |
| | Nvo-IC (ours) | 94.07 | 85.86 | 77.16 | 84.52 | 83.90 | 82.86 |
| CIFAR-10 | NODEs with Euler | 91.12 | 63.06 | 72.60 | 65.26 | 72.16 | 68.27 |
| | NODEs with Midpoint | 91.76 | 66.80 | 75.79 | 69.83 | 75.77 | 72.05 |
| | NODEs with Bosh3Solver | 92.33 | 65.97 | 75.66 | 68.51 | 75.01 | 71.28 |
| | NODEs with RK4 | 92.27 | 64.97 | 75.28 | 68.07 | 74.71 | 70.70 |
| | NFDE-IC | 91.87 | 65.75 | 76.50 | 69.97 | 76.13 | 72.09 |
| | Nvo-IC (ours) | 92.54 | 66.50 | 76.83 | 70.54 | 76.62 | 72.62 |

Table S2: Classification accuracy and robustness against stochastic noises. "Acc." denotes the classification accuracy on testing sets. "Avg." denotes the average accuracy over the four stochastic noise groups on perturbed images from two datasets. The best and the second-best results are highlighted in red and blue, respectively.

in derivative orders based on data, NvoFDE can more accurately model complex dynamic systems. This enhanced modeling capability is particularly valuable in fields such as physics, biology, and engineering, where systems often exhibit memory and non-local interactions. As with any potent modeling tool, there is a potential for misuse or unintended consequences, especially in areas directly impacting human lives, such as predictive policing or social media analytics. It is essential to consider the ethical implications and ensure that the applications of NvoFDE are aligned with societal values and norms to mitigate potential adverse effects.

# References

Anastasio, T. J. 1994. The fractional-order dynamics of brainstem vestibulo-oculomotor neurons. *Biological cybernetics*, 72(1): 69–79.

Antil, H.; Khatri, R.; Löhner, R.; and Verma, D. 2020. Fractional deep neural network via constrained optimization. *Mach. Learn.: Sci. Technol.*, 2(1): 015003.

Atangana, A.; and Baleanu, D. 2016. New fractional derivatives with nonlocal and non-singular kernel: theory and application to heat transfer model. *arXiv preprint arXiv:1602.03408*.

Caputo, M.; and Fabrizio, M. 2015. A new definition of fractional derivative without singular kernel. *Progress in Fractional Differentiation & Applications*, 1(2): 73–85.

Chamberlain, B. P.; Rowbottom, J.; Goronova, M.; Webb, S.; Rossi, E.; and Bronstein, M. M. 2021. GRAND: Graph Neural Diffusion. In *Proc. Int. Conf. Mach. Learn.*

Chami, I.; Ying, Z.; Ré, C.; and Leskovec, J. 2019. Hyperbolic graph convolutional neural networks. In *Advances Neural Inf. Process. Syst.*

Chen, R. T.; Rubanova, Y.; Bettencourt, J.; and Duvenaud, D. 2018. Neural ordinary differential equations. In *Advances Neural Inf. Process. Syst.*

Coimbra, C. F. 2003. Mechanics with variable-order differential operators. *Annalen der Physik*, 515(11-12): 692–703.

Diethelm, K. 2010. *The analysis of fractional differential equations: an application-oriented exposition using differential operators of Caputo type*, volume 2004. Springer.

Hendrycks, D.; and Dietterich, T. 2019. Benchmarking neural network robustness to common corruptions and perturbations. *arXiv preprint arXiv:1903.12261*.

Jafarian, A.; Mokhtarpour, M.; and Baleanu, D. 2017. Artificial neural network approach for a class of fractional ordinary differential equation. *Neural Computing and Applications*, 28: 765–773.

Kang, Q.; Zhao, K.; Ding, Q.; Ji, F.; Li, X.; Liang, W.; Song, Y.; and Tay, W. P. 2024a. Unleashing the Potential of Fractional Calculus in Graph Neural Networks with FROND. In *Proc. International Conference on Learning Representations*.

Kang, Q.; Zhao, K.; Song, Y.; Xie, Y.; Zhao, Y.; Wang, S.; She, R.; and Tay, W. P. 2024b. Coupling Graph Neural Networks with Fractional Order Continuous Dynamics: A Robustness Study. In *Proc. AAAI Conference on Artificial Intelligence*. Vancouver, Canada.

Kaslik, E.; and Sivasundaram, S. 2012. Nonlinear dynamics and chaos in fractional-order neural networks. *Neural networks*, 32: 245–256.

Krizhevsky, A.; and Hinton, G. 2009. Learning multiple layers of features from tiny images. *Master's thesis, Department of Computer Science, University of Toronto*.

Lasaki, F. G.; Ebrahimi, H.; and Ilie, M. 2023. A novel lagrange functional link neural network for solving variable-order fractional time-varying delay differential equations: a comparison with multilayer perceptron neural networks. *Soft Computing*, 27(17): 12595–12608.

Loshchilov, I.; and Hutter, F. 2016. Sgdr: Stochastic gradient descent with warm restarts. *arXiv preprint arXiv:1608.03983*.

Luchko, Y. 1999. Operational method in fractional calculus. *Fract. Calc. Appl. Anal*, 2(4): 463–488.

Mainardi, F. 2022. *Fractional calculus and waves in linear viscoelasticity: an introduction to mathematical models*. World Scientific.

Moghaddam, B. P.; Yaghoobi, S.; and Tenreiro Machado, J. 2016. An extended predictor–corrector algorithm for variable-order fractional delay differential equations. *Journal of Computational and Nonlinear Dynamics*, 11(6).

Munkhammar, J. 2004. Riemann-Liouville fractional derivatives and the Taylor-Riemann series.

Podlubny, I. 1999. An introduction to fractional derivatives, fractional differential equations, to methods of their solution and some of their applications. *Math. Sci. Eng*, 198(340): 0924–34008.

Rakkiyappan, R.; Udhayakumar, K.; Velmurugan, G.; Cao, J.; and Alsaedi, A. 2017. Stability and Hopf bifurcation analysis of fractional-order complex-valued neural networks with time delays. *Advances in Difference Equations*, 2017: 1–25.

Ramirez, L. E.; and Coimbra, C. F. 2011. On the variable order dynamics of the nonlinear wake caused by a sedimenting particle. *Physica D: nonlinear phenomena*, 240(13): 1111–1118.

Samko, S. G.; and Ross, B. 1993. Integration and differentiation to a variable fractional order. *Integral transforms and special functions*, 1(4): 277–300.

Sheng, H.; Sun, H.; Coopmans, C.; Chen, Y.; and Bohannan, G. 2011. A physical experimental study of variable-order fractional integrator and differentiator. *The European Physical Journal Special Topics*, 193: 93–104.

Sun, H.; Chang, A.; Zhang, Y.; and Chen, W. 2019. A review on variable-order fractional differential equations: mathematical foundations, physical models, numerical methods and applications. *Fract. Calc. Appl. Anal.*, 22(1): 27–59.

Tang, H.; Wang, D.; Huang, R.; Pei, X.; and Chen, W. 2018. A new rock creep model based on variable-order fractional derivatives and continuum damage mechanics. *Bulletin of Engineering Geology and the Environment*, 77: 375–383.

Xiao, H.; Rasul, K.; and Vollgraf, R. 2017. Fashion-mnist: a novel image dataset for benchmarking machine learning algorithms. *arXiv preprint arXiv:1708.07747*.

Xu, C.; Liao, M.; Li, P.; Guo, Y.; Xiao, Q.; and Yuan, S. 2019. Influence of multiple time delays on bifurcation of fractional-order neural networks. *Applied Mathematics and Computation*, 361: 565–582.

Yang, Y.; He, Y.; Wang, Y.; and Wu, M. 2018. Stability analysis of fractional-order neural networks: an LMI approach. *Neurocomputing*, 285: 82–93.

Yang, Z.; Liu, Y.; Bao, C.; and Shi, Z. 2020. Interpolation between residual and non-residual networks. In *International Conference on Machine Learning*, 10736–10745. PMLR.

Zúñiga-Aguilar, C.; Coronel-Escamilla, A.; Gómez-Aguilar, J.; Alvarado-Martínez, V.; and Romero-Ugalde, H. 2018. New numerical approximation for solving fractional delay differential equations of variable order using artificial neural networks. *The European Physical Journal Plus*, 133: 1–16.



% \appendix
% \input{supp.tex}

\end{document}

%% file: preamble.tex
\usepackage[T1]{fontenc}
\usepackage[utf8]{inputenc}
\usepackage{mathtools}

\usepackage{amssymb,mathrsfs}% Typical maths resource packages
\usepackage{amsthm}
\usepackage{bm}
\usepackage{scalerel}
\usepackage{nicefrac}
\usepackage{microtype} 
\usepackage[shortlabels]{enumitem}
\usepackage{graphicx}
\usepackage{epstopdf}
\DeclareGraphicsExtensions{.eps,.png,.jpg,.pdf}

\usepackage{colortbl}
\usepackage{booktabs}
\usepackage{multirow}
\usepackage{colortbl,xcolor}
\usepackage{xparse,xstring}
\usepackage{calc}
\usepackage{etoolbox}

\usepackage{array}
\newcolumntype{L}[1]{>{\raggedright\let\newline\\\arraybackslash\hspace{0pt}}m{#1}}
\newcolumntype{C}[1]{>{\centering\let\newline\\\arraybackslash\hspace{0pt}}m{#1}}
\newcolumntype{R}[1]{>{\raggedleft\let\newline\\\arraybackslash\hspace{0pt}}m{#1}}

\makeatletter
\let\MYcaption\@makecaption
\makeatother
\usepackage[font=footnotesize]{subcaption}
\makeatletter
\let\@makecaption\MYcaption
\makeatother

\usepackage{glossaries}
\makeatletter
\sfcode`\.1006

\let\oldgls\gls
\let\oldglspl\glspl

\newcommand\fussy@ifnextchar[3]{%
	\let\reserved@d=#1%
	\def\reserved@a{#2}%
	\def\reserved@b{#3}%
	\futurelet\@let@token\fussy@ifnch}
\def\fussy@ifnch{%
	\ifx\@let@token\reserved@d
		\let\reserved@c\reserved@a
	\else
		\let\reserved@c\reserved@b
	\fi
	\reserved@c}

\renewcommand{\gls}[1]{%
\oldgls{#1}\fussy@ifnextchar.{\@checkperiod}{\@}}
\renewcommand{\glspl}[1]{%
\oldglspl{#1}\fussy@ifnextchar.{\@checkperiod}{\@}}

\newcommand{\@checkperiod}[1]{%
	\ifnum\sfcode`\.=\spacefactor\else#1\fi
}

\robustify{\gls}
\robustify{\glspl}
\makeatother

\newacronym{wrt}{w.r.t.}{with respect to}
\newacronym{RHS}{R.H.S.}{right-hand side}
\newacronym{LHS}{L.H.S.}{left-hand side}
\newacronym{iid}{i.i.d.}{independent and identically distributed}
\newacronym{SOTA}{SOTA}{state-of-the-art}
\usepackage{float}

\usepackage[capitalize]{cleveref}
\crefname{equation}{}{}
\Crefname{equation}{}{}
\crefname{claim}{claim}{claims}
\crefname{step}{step}{steps}
\crefname{line}{line}{lines}
\crefname{condition}{condition}{conditions}
\crefname{dmath}{}{}
\crefname{dseries}{}{}
\crefname{dgroup}{}{}

\crefname{Problem}{Problem}{Problems}
\crefformat{Problem}{Problem~#2#1#3}
\crefrangeformat{Problem}{Problems~#3#1#4 to~#5#2#6}

\crefname{Theorem}{Theorem}{Theorems}
\crefname{Corollary}{Corollary}{Corollaries}
\crefname{Proposition}{Proposition}{Propositions}
\crefname{Lemma}{Lemma}{Lemmas}
\crefname{Definition}{Definition}{Definitions}
\crefname{Example}{Example}{Examples}
\crefname{Assumption}{Assumption}{Assumptions}
\crefname{Remark}{Remark}{Remarks}
\crefname{Rem}{Remark}{Remarks}
\crefname{remarks}{Remarks}{Remarks}
\crefname{Appendix}{Appendix}{Appendices}
\crefname{Supplement}{Supplement}{Supplements}
\crefname{Exercise}{Exercise}{Exercises}
\crefname{Theorem_A}{Theorem}{Theorems}
\crefname{Corollary_A}{Corollary}{Corollaries}
\crefname{Proposition_A}{Proposition}{Propositions}
\crefname{Lemma_A}{Lemma}{Lemmas}
\crefname{Definition_A}{Definition}{Definitions}

\usepackage{crossreftools}
\ifx\loadbreqn\undefined
	\relax
\else
	\usepackage{breqn}
\fi

\makeatletter
\def\cleartheorem#1{%
    \expandafter\let\csname#1\endcsname\relax
    \expandafter\let\csname c@#1\endcsname\relax
}
\def\clearthms#1{ \@for\tname:=#1\do{\cleartheorem\tname} }
\makeatother

\ifx\renewtheorem\undefined
	\ifx\useTheoremCounter\undefined
		\newtheorem{Theorem}{Theorem}
		\newtheorem{Corollary}{Corollary}
		\newtheorem{Proposition}{Proposition}
		
	\else

	\fi

	\newtheorem{Remark}{Remark}

\fi

\theoremstyle{remark}

\theoremstyle{plain}

\newcommand{\qednew}{\nobreak \ifvmode \relax \else
		\ifdim\lastskip<1.5em \hskip-\lastskip
			\hskip1.5em plus0em minus0.5em \fi \nobreak
		\vrule height0.75em width0.5em depth0.25em\fi}



\NewDocumentCommand{\movedownsub}{e{^_}}{%
	\IfNoValueTF{#1}{%
		\IfNoValueF{#2}{^{}}% neither ^ nor _, do nothing; if no ^ but _, add ^{}
	}{%
		^{#1}% add superscript if present
	}%
	\IfNoValueF{#2}{_{#2}}% add subscript if present
}

\let\latexchi\chi
\RenewDocumentCommand{\chi}{}{\latexchi\movedownsub}

\newcommand{\calF}{\mathcal{F}}
\newcommand{\calG}{\mathcal{G}}

\newcommand{\calV}{\mathcal{V}}

\newcommand{\bA}{\mathbf{A}}

\newcommand{\bI}{\mathbf{I}}

\newcommand{\bL}{\mathbf{L}}

\newcommand{\bV}{\mathbf{V}}

\newcommand{\bW}{\mathbf{W}}
\newcommand{\bx}{\mathbf{x}}

\newcommand{\by}{\mathbf{y}}
\newcommand{\bY}{\mathbf{Y}}

\DeclareSymbolFont{bsfletters}{OT1}{cmss}{bx}{n}
\DeclareSymbolFont{ssfletters}{OT1}{cmss}{m}{n}
\DeclareMathSymbol{\bsfGamma}{0}{bsfletters}{'000}
\DeclareMathSymbol{\ssfGamma}{0}{ssfletters}{'000}
\DeclareMathSymbol{\bsfDelta}{0}{bsfletters}{'001}
\DeclareMathSymbol{\ssfDelta}{0}{ssfletters}{'001}
\DeclareMathSymbol{\bsfTheta}{0}{bsfletters}{'002}
\DeclareMathSymbol{\ssfTheta}{0}{ssfletters}{'002}
\DeclareMathSymbol{\bsfLambda}{0}{bsfletters}{'003}
\DeclareMathSymbol{\ssfLambda}{0}{ssfletters}{'003}
\DeclareMathSymbol{\bsfXi}{0}{bsfletters}{'004}
\DeclareMathSymbol{\ssfXi}{0}{ssfletters}{'004}
\DeclareMathSymbol{\bsfPi}{0}{bsfletters}{'005}
\DeclareMathSymbol{\ssfPi}{0}{ssfletters}{'005}
\DeclareMathSymbol{\bsfSigma}{0}{bsfletters}{'006}
\DeclareMathSymbol{\ssfSigma}{0}{ssfletters}{'006}
\DeclareMathSymbol{\bsfUpsilon}{0}{bsfletters}{'007}
\DeclareMathSymbol{\ssfUpsilon}{0}{ssfletters}{'007}
\DeclareMathSymbol{\bsfPhi}{0}{bsfletters}{'010}
\DeclareMathSymbol{\ssfPhi}{0}{ssfletters}{'010}
\DeclareMathSymbol{\bsfPsi}{0}{bsfletters}{'011}
\DeclareMathSymbol{\ssfPsi}{0}{ssfletters}{'011}
\DeclareMathSymbol{\bsfOmega}{0}{bsfletters}{'012}
\DeclareMathSymbol{\ssfOmega}{0}{ssfletters}{'012}

\newcommand{\btheta}{\bm{\theta}}

\makeatletter
\newcommand*\rel@kern[1]{\kern#1\dimexpr\macc@kerna}
\newcommand*\widebar[1]{%
  \begingroup
  \def\mathaccent##1##2{%
    \rel@kern{0.8}%
    \overline{\rel@kern{-0.8}\macc@nucleus\rel@kern{0.2}}%
    \rel@kern{-0.2}%
  }%
  \macc@depth\@ne
  \let\math@bgroup\@empty \let\math@egroup\macc@set@skewchar
  \mathsurround\z@ \frozen@everymath{\mathgroup\macc@group\relax}%
  \macc@set@skewchar\relax
  \let\mathaccentV\macc@nested@a
  \macc@nested@a\relax111{#1}%
  \endgroup
}
\makeatother

\DeclareMathOperator{\var}{var}

\DeclareMathOperator{\cov}{cov}

\newcommand{\ifbcdot}[1]{\ifblank{#1}{\cdot}{#1}}

\DeclarePairedDelimiterX\abs[1]{\lvert}{\rvert}{\ifbcdot{#1}}
\DeclarePairedDelimiterX\parens[1]{(}{)}{\ifbcdot{#1}}
\DeclarePairedDelimiterX\brk[1]{[}{]}{\ifbcdot{#1}}
\DeclarePairedDelimiterX\braces[1]{\{}{\}}{\ifbcdot{#1}}
\DeclarePairedDelimiterX\angles[1]{\langle}{\rangle}{\ifblank{#1}{\cdot,\cdot}{#1}}
\DeclarePairedDelimiterX\ip[2]{\langle}{\rangle}{\ifbcdot{#1},\ifbcdot{#2}}
\DeclarePairedDelimiterX\norm[1]{\lVert}{\rVert}{\ifbcdot{#1}}
\DeclarePairedDelimiterX\ceil[1]{\lceil}{\rceil}{\ifbcdot{#1}}
\DeclarePairedDelimiterX\floor[1]{\lfloor}{\rfloor}{\ifbcdot{#1}}

\DeclareFontFamily{U}{matha}{\hyphenchar\font45}
\DeclareFontShape{U}{matha}{m}{n}{
      <5> <6> <7> <8> <9> <10> gen * matha
      <10.95> matha10 <12> <14.4> <17.28> <20.74> <24.88> matha12
      }{}
\DeclareSymbolFont{matha}{U}{matha}{m}{n}
\DeclareFontSubstitution{U}{matha}{m}{n}

\DeclareFontFamily{U}{mathx}{\hyphenchar\font45}
\DeclareFontShape{U}{mathx}{m}{n}{
      <5> <6> <7> <8> <9> <10>
      <10.95> <12> <14.4> <17.28> <20.74> <24.88>
      mathx10
      }{}
\DeclareSymbolFont{mathx}{U}{mathx}{m}{n}
\DeclareFontSubstitution{U}{mathx}{m}{n}

\DeclareMathDelimiter{\vvvert}{0}{matha}{"7E}{mathx}{"17}
\DeclarePairedDelimiterX\vertiii[1]{\vvvert}{\vvvert}{\ifbcdot{#1}}

\DeclarePairedDelimiterXPP\trace[1]{\operatorname{Tr}}{(}{)}{}{\ifbcdot{#1}} % column vector
\DeclarePairedDelimiterXPP\col[1]{\operatorname{col}}{\{}{\}}{}{\ifbcdot{#1}} % column vector
\DeclarePairedDelimiterXPP\row[1]{\operatorname{row}}{\{}{\}}{}{\ifbcdot{#1}} % row vector
\DeclarePairedDelimiterXPP\erf[1]{\operatorname{erf}}{(}{)}{}{\ifbcdot{#1}}
\DeclarePairedDelimiterXPP\erfc[1]{\operatorname{erfc}}{(}{)}{}{\ifbcdot{#1}}
\DeclarePairedDelimiterXPP\KLD[2]{D}{(}{)}{}{\ifbcdot{#1}\, \delimsize\|\, \ifbcdot{#2}} % KL divergence
\DeclarePairedDelimiterXPP\op[2]{\operatorname{#1}}{(}{)}{}{#2} % general operator

\newcommand{\T}{^{\mkern-1.5mu\mathop\intercal}}% transpose notation
\newcommand{\ud}{\,\mathrm{d}} % for integrals like \int f(x) \ud x
\DeclarePairedDelimiterXPP\indicate[1]{{\bf 1}}{\{}{\}}{}{\ifbcdot{#1}}

\NewDocumentCommand\ofrac{s m}{%
	\IfBooleanTF#1%
	{\dfrac{1}{#2}}%
	{\frac{1}{#2}}%
}
\NewDocumentCommand\ddfrac{s m m}{%
	\IfBooleanTF#1%
	{\dfrac{\mathrm{d} {#2}}{\mathrm{d} {#3}}}%
	{\frac{\mathrm{d} {#2}}{\mathrm{d} {#3}}}%
}
\NewDocumentCommand\ppfrac{s m m}{%
	\IfBooleanTF#1%
	{\dfrac{\partial {#2}}{\partial {#3}}}%
	{\frac{\partial {#2}}{\partial {#3}}}%
}

\providecommand\given{}
\DeclarePairedDelimiterX\Set[2]\{\}{%
\renewcommand\given{\SetSymbol[\delimsize]{#1}}
#2
}
\DeclarePairedDelimiterX\Setc[1]\{\}{%
\renewcommand\given{\SetSymbol{:}}
#1
}

\NewDocumentCommand\set{s o m}{%
	\IfBooleanTF#1%
	{\IfValueTF{#2}{\Set*{#2}{#3}}{\Setc*{#3}}}%
	{\IfValueTF{#2}{\Set{#2}{#3}}{\Setc{#3}}}%
}

\NewDocumentCommand{\evalat}{ s O{\big} m e{_^} }{%
\IfBooleanTF{#1}%
{\left. #3 \right|}{#3#2|}%
\IfValueT{#4}{_{#4}}%
\IfValueT{#5}{^{#5}}%
}

\providecommand\given{}
\DeclarePairedDelimiterXPP\cprob[1]{}(){}{
\renewcommand\given{\nonscript\,\delimsize\vert\allowbreak\nonscript\,\mathopen{}}%
#1%
}
\DeclarePairedDelimiterXPP\cexp[1]{}[]{}{
\renewcommand\given{\nonscript\,\delimsize\vert\allowbreak\nonscript\,\mathopen{}}%
#1%
}

\DeclareDocumentCommand \P { s e{_^} d() g } {%
	\mathbb{P}%
	\IfBooleanTF{#1}%
		{
			\IfValueT{#2}{_{#2}}%
			\IfValueT{#3}{^{#3}}%
			\IfValueTF{#5}{\cprob{#4 \given #5}}{\IfValueT{#4}{\cprob{#4}}}%
		}%
		{
			\IfValueT{#2}{_{#2}}%
			\IfValueT{#3}{^{#3}}%
			\IfValueTF{#5}{\cprob*{#4 \given #5}}{\IfValueT{#4}{\cprob*{#4}}}%
		}%
}

\DeclareDocumentCommand \E { s e{_^} o g } {%
	\mathbb{E}%
	\IfBooleanTF{#1}%
		{
			\IfValueT{#2}{_{#2}}%
			\IfValueT{#3}{^{#3}}%
			\IfValueTF{#5}{\cexp{#4 \given #5}}{\IfValueT{#4}{\cexp{#4}}}%
		}%
		{
			\IfValueT{#2}{_{#2}}%
			\IfValueT{#3}{^{#3}}%	
			\IfValueTF{#5}{\cexp*{#4 \given #5}}{\IfValueT{#4}{\cexp*{#4}}}%		
		}%
}

\DeclareDocumentCommand \Var { s e{_^} d() g } {%
	\var%
	\IfBooleanTF{#1}%
		{
			\IfValueT{#2}{_{#2}}%
			\IfValueT{#3}{^{#3}}%
			\IfValueTF{#5}{\cprob{#4 \given #5}}{\IfValueT{#4}{\cprob{#4}}}%
		}%
		{
			\IfValueT{#2}{_{#2}}%
			\IfValueT{#3}{^{#3}}%	
			\IfValueTF{#5}{\cprob*{#4 \given #5}}{\IfValueT{#4}{\cprob*{#4}}}%		
		}%
}

\DeclareDocumentCommand \Cov { s e{_^} d() g } {%
	\cov%
	\IfBooleanTF{#1}%
		{
			\IfValueT{#2}{_{#2}}%
			\IfValueT{#3}{^{#3}}%
			\IfValueTF{#5}{\cprob{#4 \given #5}}{\IfValueT{#4}{\cprob{#4}}}%
		}%
		{
			\IfValueT{#2}{_{#2}}%
			\IfValueT{#3}{^{#3}}%	
			\IfValueTF{#5}{\cprob*{#4 \given #5}}{\IfValueT{#4}{\cprob*{#4}}}%		
		}%
}

\ExplSyntaxOn
\NewDocumentCommand \dist {m o o} {%
\mathrm{#1}\left(%
	\IfValueT{#3}{%
		\tl_if_blank:nTF{ #3 }{\cdot\, \middle|\, }{#3\, \middle|\, }%
	}
	\IfValueT{#2}{#2}%
\right)%
}
\ExplSyntaxOff

\NewDocumentCommand {\cbrace} {t+ D[]{black} D(){\widthof{#5}} m m } {%
	\begingroup%
		\color{#2}
		\IfBooleanTF{#1}{%
			\overbrace{#4}^%
		}{
			\underbrace{#4}_%
		}%
		{\parbox[c]{#3}{\centering\footnotesize{#5}}}%
	\endgroup% 
}

\let\oldforall\forall
\renewcommand{\forall}{\oldforall \, }

\let\oldexist\exists
\renewcommand{\exists}{\oldexist \, }

\makeatletter

\newcommand{\rankcolor}[2]{%
	\expandafter\renewcommand\csname #1\endcsname[1]{%
		\ifblank{##1}{%
			{\color{#2} \textbf{#2}}%
		}{%
			\ifmmode
				\textcolor{#2}{\bm{##1}}%
			\else%
				{\color{#2} \textbf{##1}}%
			\fi	
		}%
	}
}

\providecommand{\first}{}
\providecommand{\second}{}

\rankcolor{first}{red}
\rankcolor{second}{blue}
\rankcolor{third}{cyan}
\makeatother

\graphicspath{{./Figures/}{./figures/}}
\DeclareDocumentCommand{\includeCroppedPdf}{ o O{./Figures/} m }{
	\IfFileExists{#2#3-crop.pdf}{}{%
		\immediate\write18{pdfcrop #2#3.pdf #2#3-crop.pdf}}%
	\includegraphics[#1]{#2#3-crop.pdf}
}

\makeatletter
\newcommand*{\addFileDependency}[1]{% argument=file name and extension
  \typeout{(#1)}
  \@addtofilelist{#1}
  \IfFileExists{#1}{}{\typeout{No file #1.}}
}
\makeatother

\definecolor{gray90}{gray}{0.9}
\def\colorlist{red,blue,brown,cyan,darkgray,gray,lightgray,green,lime,magenta,olive,orange,pink,purple,teal,violet,white,yellow}

\makeatletter
\def\startcomment{[}
\ifx\nohighlights\undefined
	\newcommand{\createcolor}[1]{%
			\expandafter\newcommand\csname #1\endcsname[1]{{\color{#1} ##1}}%
	}
	\newcommand{\msout}[1]{\text{\color{green} \sout{\ensuremath{#1}}}}
	\newcommand{\del}[1]{{\color{green}\ifmmode \msout{#1}\else\sout{#1}\fi}}
\else
	\newcommand{\createcolor}[1]{%
			\expandafter\newcommand\csname #1\endcsname[1]{%
				\noexpandarg%
				\StrChar{##1}{1}[\firstletter]%
				\if\firstletter\startcomment%
					\relax
				\else%
					##1
				\fi
			}%
	}
	\newcommand{\msout}[1]{}
	\newcommand{\del}[1]{}
\fi

\def\@tempa#1,{%
    \ifx\relax#1\relax\else
        \createcolor{#1}%
        \expandafter\@tempa
    \fi
}
\expandafter\@tempa\colorlist,\relax,
\makeatother

\newcommand{\hhide}[1]{}
\ifx\diagnoselabel\undefined
	\relax
\else
	\makeatletter
	\def\@testdef #1#2#3{%
		\def\reserved@a{#3}\expandafter \ifx \csname #1@#2\endcsname
			\reserved@a  \else
			\typeout{^^Jlabel #2 changed:^^J%
				\meaning\reserved@a^^J%
				\expandafter\meaning\csname #1@#2\endcsname^^J}%
			\@tempswatrue \fi}
	\makeatother
\fi

%% file: main.bbl
\begin{thebibliography}{62}
\providecommand{\natexlab}[1]{#1}

\bibitem[{Abadi et~al.(2016)Abadi, Barham, Chen, Chen, Davis, Dean, Devin,
  Ghemawat, Irving, Isard et~al.}]{abadi2016tensorflow}
Abadi, M.; Barham, P.; Chen, J.; Chen, Z.; Davis, A.; Dean, J.; Devin, M.;
  Ghemawat, S.; Irving, G.; Isard, M.; et~al. 2016.
\newblock $\{$TensorFlow$\}$: a system for $\{$Large-Scale$\}$ machine
  learning.
\newblock In \emph{12th USENIX symposium on operating systems design and
  implementation (OSDI 16)}, 265--283.

\bibitem[{Akg{\"u}l, Baleanu et~al.(2017)}]{akgul2017solutions}
Akg{\"u}l, A.; Baleanu, D.; et~al. 2017.
\newblock On solutions of variable-order fractional differential equations.
\newblock \emph{An International Journal of Optimization and Control: Theories
  \& Applications (IJOCTA)}, 7(1): 112--116.

\bibitem[{Anastasio(1994)}]{anastasio1994fractional}
Anastasio, T.~J. 1994.
\newblock The fractional-order dynamics of brainstem vestibulo-oculomotor
  neurons.
\newblock \emph{Biological cybernetics}, 72(1): 69--79.

\bibitem[{Antil et~al.(2020)Antil, Khatri, L{\"o}hner, and
  Verma}]{antil2020fractional}
Antil, H.; Khatri, R.; L{\"o}hner, R.; and Verma, D. 2020.
\newblock Fractional deep neural network via constrained optimization.
\newblock \emph{Mach. Learn.: Sci. Technol.}, 2(1): 015003.

\bibitem[{Bodnar et~al.(2022)Bodnar, Giovanni, Chamberlain, Li{\`o}, and
  Bronstein}]{crifraben:sheaf2022}
Bodnar, C.; Giovanni, F.~D.; Chamberlain, B.~P.; Li{\`o}, P.; and Bronstein,
  M.~M. 2022.
\newblock Neural Sheaf Diffusion: A Topological Perspective on Heterophily and
  Oversmoothing in {GNN}s.
\newblock In \emph{Advances Neural Inf. Process. Syst.}

\bibitem[{Chamberlain et~al.(2021{\natexlab{a}})Chamberlain, Rowbottom,
  Gorinova, Bronstein, Webb, and Rossi}]{chamrowgor:grand2021}
Chamberlain, B.; Rowbottom, J.; Gorinova, M.~I.; Bronstein, M.; Webb, S.; and
  Rossi, E. 2021{\natexlab{a}}.
\newblock Grand: Graph neural diffusion.
\newblock In \emph{Proc. Int. Conf. Mach. Learn.}, 1407--1418.

\bibitem[{Chamberlain et~al.(2021{\natexlab{b}})Chamberlain, Rowbottom,
  Goronova, Webb, Rossi, and Bronstein}]{chamberlain2021grand}
Chamberlain, B.~P.; Rowbottom, J.; Goronova, M.; Webb, S.; Rossi, E.; and
  Bronstein, M.~M. 2021{\natexlab{b}}.
\newblock GRAND: Graph Neural Diffusion.
\newblock In \emph{Proc. Int. Conf. Mach. Learn.}

\bibitem[{Chami et~al.(2019)Chami, Ying, R{\'e}, and
  Leskovec}]{chami2019hyperbolic_GCNN}
Chami, I.; Ying, Z.; R{\'e}, C.; and Leskovec, J. 2019.
\newblock Hyperbolic graph convolutional neural networks.
\newblock In \emph{Advances Neural Inf. Process. Syst.}

\bibitem[{Chen et~al.(2018)Chen, Rubanova, Bettencourt, and
  Duvenaud}]{chen2018neural}
Chen, R.~T.; Rubanova, Y.; Bettencourt, J.; and Duvenaud, D. 2018.
\newblock Neural ordinary differential equations.
\newblock In \emph{Advances Neural Inf. Process. Syst.}

\bibitem[{Coimbra(2003)}]{coimbra2003mechanics}
Coimbra, C.~F. 2003.
\newblock Mechanics with variable-order differential operators.
\newblock \emph{Annalen der Physik}, 515(11-12): 692--703.

\bibitem[{Coleman and Noll(1961)}]{coleman1961foundations}
Coleman, B.~D.; and Noll, W. 1961.
\newblock Foundations of linear viscoelasticity.
\newblock \emph{Rev. Modern Phys.}, 33(2): 239.

\bibitem[{Cui et~al.(2023)Cui, Zhang, Chu, Hu, and Li}]{cui2023robustness}
Cui, W.; Zhang, H.; Chu, H.; Hu, P.; and Li, Y. 2023.
\newblock On robustness of neural ODEs image classifiers.
\newblock \emph{Information Sciences}, 632: 576--593.

\bibitem[{Dai et~al.(2024)Dai, Qu, Chen, Zhang, and Xu}]{dai2024recode}
Dai, S.; Qu, C.; Chen, S.; Zhang, X.; and Xu, J. 2024.
\newblock Recode: Modeling repeat consumption with neural ode.
\newblock In \emph{Proc. Int. ACM SIGIR Conference on Research and Development
  in Information Retrieval}, 2599--2603.

\bibitem[{Diethelm(2010)}]{diethelm2010analysis}
Diethelm, K. 2010.
\newblock \emph{The analysis of fractional differential equations: an
  application-oriented exposition using differential operators of Caputo type},
  volume 2004.
\newblock Springer.

\bibitem[{Diethelm, Ford, and Freed(2004)}]{diethelm2004detailed}
Diethelm, K.; Ford, N.~J.; and Freed, A.~D. 2004.
\newblock Detailed error analysis for a fractional Adams method.
\newblock \emph{Numer. Algorithms}, 36: 31--52.

\bibitem[{Dupont, Doucet, and Teh(2019)}]{dupont2019augmented}
Dupont, E.; Doucet, A.; and Teh, Y.~W. 2019.
\newblock Augmented neural odes.
\newblock In \emph{Advances Neural Inf. Process. Syst.}, 1--11.

\bibitem[{Gl{\"o}ckle and Nonnenmacher(1995)}]{glockle1995fractional}
Gl{\"o}ckle, W.~G.; and Nonnenmacher, T.~F. 1995.
\newblock A fractional calculus approach to self-similar protein dynamics.
\newblock \emph{Biophysical Journal}, 68(1): 46--53.

\bibitem[{Gravina, Bacciu, and Gallicchio(2022)}]{gravina2022anti}
Gravina, A.; Bacciu, D.; and Gallicchio, C. 2022.
\newblock Anti-symmetric dgn: A stable architecture for deep graph networks.
\newblock In \emph{Proc. Int. Conf. Learn. Representations}.

\bibitem[{Haber and Ruthotto(2017)}]{haber2017stable}
Haber, E.; and Ruthotto, L. 2017.
\newblock Stable architectures for deep neural networks.
\newblock \emph{Inverse Problems}, 34(1): 1--23.

\bibitem[{Kang et~al.(2025)Kang, Li, Zhao, Cui, Zhao, Deng, and
  Tay}]{KanLiZha:C25}
Kang, Q.; Li, X.; Zhao, K.; Cui, W.; Zhao, Y.; Deng, W.; and Tay, W.~P. 2025.
\newblock Efficient Training of Neural Fractional-Order Differential Equation
  via Adjoint Backpropagation.
\newblock In \emph{Proc. AAAI Conference on Artificial Intelligence}. USA.

\bibitem[{Kang et~al.(2021)Kang, Song, Ding, and Tay}]{kang2021Neurips}
Kang, Q.; Song, Y.; Ding, Q.; and Tay, W.~P. 2021.
\newblock Stable neural {ODE} with {Lyapunov-stable} equilibrium points for
  defending against adversarial attacks.
\newblock In \emph{Advances Neural Inf. Process. Syst.}

\bibitem[{Kang et~al.(2024{\natexlab{a}})Kang, Zhao, Ding, Ji, Li, Liang, Song,
  and Tay}]{KanZhaDin:C24}
Kang, Q.; Zhao, K.; Ding, Q.; Ji, F.; Li, X.; Liang, W.; Song, Y.; and Tay,
  W.~P. 2024{\natexlab{a}}.
\newblock Unleashing the Potential of Fractional Calculus in Graph Neural
  Networks with {FROND}.
\newblock In \emph{Proc. International Conference on Learning Representations}.

\bibitem[{Kang et~al.(2023)Kang, Zhao, Song, Wang, and Tay}]{KanZhaSon:C23}
Kang, Q.; Zhao, K.; Song, Y.; Wang, S.; and Tay, W.~P. 2023.
\newblock Node Embedding from Neural {Hamiltonian} Orbits in Graph Neural
  Networks.
\newblock In \emph{Proc. International Conference on Machine Learning},
  15786--15808.

\bibitem[{Kang et~al.(2024{\natexlab{b}})Kang, Zhao, Song, Xie, Zhao, Wang,
  She, and Tay}]{ZhaKanSon:C24}
Kang, Q.; Zhao, K.; Song, Y.; Xie, Y.; Zhao, Y.; Wang, S.; She, R.; and Tay,
  W.~P. 2024{\natexlab{b}}.
\newblock Coupling Graph Neural Networks with Fractional Order Continuous
  Dynamics: {A} Robustness Study.
\newblock In \emph{Proc. AAAI Conference on Artificial Intelligence}.
  Vancouver, Canada.

\bibitem[{Kaslik and Sivasundaram(2012)}]{kaslik2012nonlinear}
Kaslik, E.; and Sivasundaram, S. 2012.
\newblock Nonlinear dynamics and chaos in fractional-order neural networks.
\newblock \emph{Neural networks}, 32: 245--256.

\bibitem[{Kingma and Ba(2014)}]{kingma2014adam}
Kingma, D.~P.; and Ba, J. 2014.
\newblock Adam: A method for stochastic optimization.
\newblock \emph{arXiv preprint arXiv:1412.6980}.

\bibitem[{Kipf and Welling(2017)}]{kipf2017semi}
Kipf, T.~N.; and Welling, M. 2017.
\newblock Semi-Supervised Classification with Graph Convolutional Networks.
\newblock In \emph{Proc. Int. Conf. Learn. Representations}.

\bibitem[{Krizhevsky and Hinton(2009)}]{KriTR2009}
Krizhevsky, A.; and Hinton, G. 2009.
\newblock Learning multiple layers of features from tiny images.
\newblock \emph{Master's thesis, Department of Computer Science, University of
  Toronto}.

\bibitem[{Lasaki, Ebrahimi, and Ilie(2023)}]{lasaki2023novel}
Lasaki, F.~G.; Ebrahimi, H.; and Ilie, M. 2023.
\newblock A novel lagrange functional link neural network for solving
  variable-order fractional time-varying delay differential equations: a
  comparison with multilayer perceptron neural networks.
\newblock \emph{Soft Computing}, 27(17): 12595--12608.

\bibitem[{Liu et~al.(2022)Liu, Wang, Luo, and Luo}]{liu2022regularized}
Liu, Z.; Wang, Y.; Luo, Y.; and Luo, C. 2022.
\newblock A Regularized Graph Neural Network Based on Approximate Fractional
  Order Gradients.
\newblock \emph{Mathematics}, 10(8): 1320.

\bibitem[{Lorenzo and Hartley(2002)}]{lorenzo2002variable}
Lorenzo, C.~F.; and Hartley, T.~T. 2002.
\newblock Variable order and distributed order fractional operators.
\newblock \emph{Nonlinear dynamics}, 29: 57--98.

\bibitem[{Machado, Kiryakova, and Mainardi(2011)}]{machado2011recent}
Machado, J.~T.; Kiryakova, V.; and Mainardi, F. 2011.
\newblock Recent history of fractional calculus.
\newblock \emph{Communications in nonlinear science and numerical simulation},
  16(3): 1140--1153.

\bibitem[{Meng et~al.(2016)Meng, Yin, Zhou, and Wu}]{meng2016fractional}
Meng, R.; Yin, D.; Zhou, C.; and Wu, H. 2016.
\newblock Fractional description of time-dependent mechanical property
  evolution in materials with strain softening behavior.
\newblock \emph{Applied Mathematical Modelling}, 40(1): 398--406.

\bibitem[{Moghaddam, Yaghoobi, and
  Tenreiro~Machado(2016)}]{moghaddam2016extended}
Moghaddam, B.~P.; Yaghoobi, S.; and Tenreiro~Machado, J. 2016.
\newblock An extended predictor--corrector algorithm for variable-order
  fractional delay differential equations.
\newblock \emph{Journal of Computational and Nonlinear Dynamics}, 11(6).

\bibitem[{Nigmatullin(1986)}]{nigmatullin1986realization}
Nigmatullin, R. 1986.
\newblock The realization of the generalized transfer equation in a medium with
  fractal geometry.
\newblock \emph{Physica status solidi (b)}, 133(1): 425--430.

\bibitem[{Nobis et~al.(2023)Nobis, Aversa, Springenberg, Detzel, Ermon,
  Nakajima, Murray-Smith, Lapuschkin, Knochenhauer, Oala
  et~al.}]{nobis2023generative}
Nobis, G.; Aversa, M.; Springenberg, M.; Detzel, M.; Ermon, S.; Nakajima, S.;
  Murray-Smith, R.; Lapuschkin, S.; Knochenhauer, C.; Oala, L.; et~al. 2023.
\newblock Generative Fractional Diffusion Models.
\newblock \emph{arXiv preprint arXiv:2310.17638}.

\bibitem[{Obembe, Hossain, and Abu-Khamsin(2017)}]{obembe2017variable}
Obembe, A.~D.; Hossain, M.~E.; and Abu-Khamsin, S.~A. 2017.
\newblock Variable-order derivative time fractional diffusion model for
  heterogeneous porous media.
\newblock \emph{Journal of Petroleum Science and Engineering}, 152: 391--405.

\bibitem[{OpenAI(2022)}]{openai2022chatgpt4}
OpenAI. 2022.
\newblock ChatGPT-4.
\newblock Available at: \url{https://www.openai.com} (Accessed: 10 April 2024).

\bibitem[{Paszke et~al.(2019)Paszke, Gross, Massa, Lerer, Bradbury, Chanan,
  Killeen, Lin, Gimelshein, Antiga et~al.}]{paszke2019pytorch}
Paszke, A.; Gross, S.; Massa, F.; Lerer, A.; Bradbury, J.; Chanan, G.; Killeen,
  T.; Lin, Z.; Gimelshein, N.; Antiga, L.; et~al. 2019.
\newblock Pytorch: An imperative style, high-performance deep learning library.
\newblock \emph{Advances Neural Inf. Process. Syst.}

\bibitem[{Pfau et~al.(2020)Pfau, Spencer, Matthews, and Foulkes}]{pfau2020ab}
Pfau, D.; Spencer, J.~S.; Matthews, A.~G.; and Foulkes, W. M.~C. 2020.
\newblock Ab initio solution of the many-electron Schr{\"o}dinger equation with
  deep neural networks.
\newblock \emph{Physical review research}, 2(3): 033429.

\bibitem[{Platonov et~al.(2023)Platonov, Kuznedelev, Diskin, Babenko, and
  Prokhorenkova}]{platonov2023critical}
Platonov, O.; Kuznedelev, D.; Diskin, M.; Babenko, A.; and Prokhorenkova, L.
  2023.
\newblock A critical look at the evaluation of GNNs under heterophily: Are we
  really making progress?
\newblock \emph{arXiv preprint arXiv:2302.11640}.

\bibitem[{Raissi, Perdikaris, and Karniadakis(2019)}]{raissi2019physics}
Raissi, M.; Perdikaris, P.; and Karniadakis, G.~E. 2019.
\newblock Physics-informed neural networks: A deep learning framework for
  solving forward and inverse problems involving nonlinear partial differential
  equations.
\newblock \emph{J. Comput. Phys.}, 378: 686--707.

\bibitem[{Samko and Ross(1993)}]{samko1993integration}
Samko, S.~G.; and Ross, B. 1993.
\newblock Integration and differentiation to a variable fractional order.
\newblock \emph{Integral transforms and special functions}, 1(4): 277--300.

\bibitem[{Scalas, Gorenflo, and Mainardi(2000)}]{scalas2000fractional}
Scalas, E.; Gorenflo, R.; and Mainardi, F. 2000.
\newblock Fractional calculus and continuous-time finance.
\newblock \emph{Physica A: Statistical Mechanics and its Applications},
  284(1-4): 376--384.

\bibitem[{Shukla et~al.(2020)Shukla, Di~Leoni, Blackshire, Sparkman, and
  Karniadakis}]{shukla2020physics}
Shukla, K.; Di~Leoni, P.~C.; Blackshire, J.; Sparkman, D.; and Karniadakis,
  G.~E. 2020.
\newblock Physics-informed neural network for ultrasound nondestructive
  quantification of surface breaking cracks.
\newblock \emph{Journal of Nondestructive Evaluation}, 39: 1--20.

\bibitem[{Song et~al.(2022)Song, Kang, Wang, Zhao, and Tay}]{SonKanWan:C22}
Song, Y.; Kang, Q.; Wang, S.; Zhao, K.; and Tay, W.~P. 2022.
\newblock On the Robustness of Graph Neural Diffusion to Topology
  Perturbations.
\newblock In \emph{Advances Neural Inf. Process. Syst.}

\bibitem[{Song et~al.(2021)Song, Sohl-Dickstein, Kingma, Kumar, Ermon, and
  Poole}]{songscore}
Song, Y.; Sohl-Dickstein, J.; Kingma, D.~P.; Kumar, A.; Ermon, S.; and Poole,
  B. 2021.
\newblock Score-Based Generative Modeling through Stochastic Differential
  Equations.
\newblock In \emph{International Conference on Learning Representations}.

\bibitem[{Sun et~al.(2019)Sun, Chang, Zhang, and Chen}]{sun2019review}
Sun, H.; Chang, A.; Zhang, Y.; and Chen, W. 2019.
\newblock A review on variable-order fractional differential equations:
  mathematical foundations, physical models, numerical methods and
  applications.
\newblock \emph{Fract. Calc. Appl. Anal.}, 22(1): 27--59.

\bibitem[{Sun et~al.(2011)Sun, Chen, Wei, and Chen}]{sun2011comparative}
Sun, H.; Chen, W.; Wei, H.; and Chen, Y. 2011.
\newblock A comparative study of constant-order and variable-order fractional
  models in characterizing memory property of systems.
\newblock \emph{The european physical journal special topics}, 193(1):
  185--192.

\bibitem[{Vaswani et~al.(2017)Vaswani, Shazeer, Parmar, Uszkoreit, Jones,
  Gomez, Kaiser, and Polosukhin}]{vaswani2017attention}
Vaswani, A.; Shazeer, N.; Parmar, N.; Uszkoreit, J.; Jones, L.; Gomez, A.~N.;
  Kaiser, {\L}.; and Polosukhin, I. 2017.
\newblock Attention is all you need.
\newblock In \emph{Advances Neural Inf. Process. Syst.}

\bibitem[{Veli{\v{c}}kovi{\'{c}} et~al.(2018)Veli{\v{c}}kovi{\'{c}}, Cucurull,
  Casanova, Romero, Li{\`{o}}, and Bengio}]{velickovic2018graph}
Veli{\v{c}}kovi{\'{c}}, P.; Cucurull, G.; Casanova, A.; Romero, A.; Li{\`{o}},
  P.; and Bengio, Y. 2018.
\newblock Graph Attention Networks.
\newblock In \emph{Proc. Int. Conf. Learn. Representations}, 1--12.

\bibitem[{Weinan(2017)}]{weinan2017proposal}
Weinan, E. 2017.
\newblock A proposal on machine learning via dynamical systems.
\newblock \emph{Commun. Math. Statist.}, 1(5): 1--11.

\bibitem[{Xiao, Rasul, and Vollgraf(2017)}]{xiao2017fashion}
Xiao, H.; Rasul, K.; and Vollgraf, R. 2017.
\newblock Fashion-mnist: a novel image dataset for benchmarking machine
  learning algorithms.
\newblock \emph{arXiv preprint arXiv:1708.07747}.

\bibitem[{Xu et~al.(2019)Xu, Liao, Li, Guo, Xiao, and Yuan}]{xu2019influence}
Xu, C.; Liao, M.; Li, P.; Guo, Y.; Xiao, Q.; and Yuan, S. 2019.
\newblock Influence of multiple time delays on bifurcation of fractional-order
  neural networks.
\newblock \emph{Applied Mathematics and Computation}, 361: 565--582.

\bibitem[{Xu et~al.(2022)Xu, Chen, Li, and Duvenaud}]{xu2022infinitely}
Xu, W.; Chen, R.~T.; Li, X.; and Duvenaud, D. 2022.
\newblock Infinitely deep bayesian neural networks with stochastic differential
  equations.
\newblock In \emph{International Conference on Artificial Intelligence and
  Statistics}, 721--738. PMLR.

\bibitem[{Yan et~al.(2018)Yan, Du, Tan, and Feng}]{yan2019robustness}
Yan, H.; Du, J.; Tan, V.~Y.; and Feng, J. 2018.
\newblock On robustness of neural ordinary differential equations.
\newblock In \emph{Advances Neural Inf. Process. Syst.}, 1--13.

\bibitem[{Yang et~al.(2018)Yang, He, Wang, and Wu}]{yang2018stability}
Yang, Y.; He, Y.; Wang, Y.; and Wu, M. 2018.
\newblock Stability analysis of fractional-order neural networks: an LMI
  approach.
\newblock \emph{Neurocomputing}, 285: 82--93.

\bibitem[{Zhang et~al.(2018)Zhang, Han, Wang, Car, and E}]{zhang2018deep}
Zhang, L.; Han, J.; Wang, H.; Car, R.; and E, W. 2018.
\newblock Deep potential molecular dynamics: a scalable model with the accuracy
  of quantum mechanics.
\newblock \emph{Physical review letters}, 120(14): 143001.

\bibitem[{Zhao et~al.(2023)Zhao, Kang, Song, She, Wang, and
  Tay}]{ZhaKanSon:C23}
Zhao, K.; Kang, Q.; Song, Y.; She, R.; Wang, S.; and Tay, W.~P. 2023.
\newblock Graph Neural Convection-Diffusion with Heterophily.
\newblock In \emph{Proc. Inter. Joint Conf. Artificial Intell.} China.

\bibitem[{Zhu et~al.(2020)Zhu, Pan, Zhou, Wu, Cao, and Wang}]{zhu2020GIL}
Zhu, S.; Pan, S.; Zhou, C.; Wu, J.; Cao, Y.; and Wang, B. 2020.
\newblock Graph Geometry Interaction Learning.
\newblock In \emph{Advances Neural Inf. Process. Syst.}

\bibitem[{Zhu et~al.(2021)Zhu, Xu, Darve, and Beroza}]{zhu2021general}
Zhu, W.; Xu, K.; Darve, E.; and Beroza, G.~C. 2021.
\newblock A general approach to seismic inversion with automatic
  differentiation.
\newblock \emph{Computers \& Geosciences}, 151: 104751.

\bibitem[{Zhu et~al.(2019)Zhu, Zabaras, Koutsourelakis, and
  Perdikaris}]{zhu2019physics}
Zhu, Y.; Zabaras, N.; Koutsourelakis, P.-S.; and Perdikaris, P. 2019.
\newblock Physics-constrained deep learning for high-dimensional surrogate
  modeling and uncertainty quantification without labeled data.
\newblock \emph{J. Comput. Phys.}, 394: 56--81.

\end{thebibliography}
